\documentclass[10pt]{article}

\usepackage[utf8]{inputenc}
\usepackage[T1]{fontenc}
\usepackage[letterpaper,textwidth=5.5in,textheight=9in]{geometry}
\usepackage{mathptmx}
\usepackage{amsmath,amssymb}
\usepackage{microtype}
\usepackage{graphicx}
\usepackage{wrapfig}
\usepackage{booktabs}
\usepackage{placeins}
\usepackage{algorithm}
\usepackage{algpseudocode}
\usepackage{titlesec}
\usepackage{titling}
\usepackage{xcolor}
\usepackage[round,sort&compress]{natbib}
\usepackage{hyperref}

\definecolor{arxivlink}{rgb}{0.10,0.20,0.55}
\definecolor{tblgain}{rgb}{0.00,0.45,0.10}
\definecolor{tblloss}{rgb}{0.72,0.04,0.04}
\hypersetup{colorlinks=true,linkcolor=arxivlink,citecolor=arxivlink,urlcolor=arxivlink}


\titleformat{\section}{\large\bfseries}{\thesection}{0.75em}{}
\titleformat{\subsection}{\normalsize\bfseries}{\thesubsection}{0.75em}{}
\titlespacing*{\section}{0pt}{2.2ex plus 1ex minus .2ex}{1.0ex plus .2ex}
\titlespacing*{\subsection}{0pt}{1.6ex plus 1ex minus .2ex}{0.6ex plus .2ex}

\pretitle{\begin{center}\LARGE\bfseries}
\posttitle{\par\end{center}\vskip 0.4em}
\preauthor{\begin{center}\large}
\postauthor{\par\end{center}}
\predate{}
\postdate{}

\renewenvironment{abstract}{\begin{center}\large\bfseries Abstract\end{center}\vspace{-0.5ex}\begin{quote}}{\end{quote}\vspace{1.0ex}}

\newcommand{\E}{\mathbb{E}}
\newcommand{\KL}[2]{D_{\mathrm{KL}}\!\left(#1 \,\big\|\, #2\right)}
\newcommand{\pibase}{\pi_{\mathrm{base}}}
\newcommand{\pith}{\pi_{\theta}}
\newcommand{\piold}{\pi_{\theta_{\mathrm{old}}}}
\newcommand{\wdl}{\mathbf{p}}
\DeclareMathOperator{\Ent}{H}
\DeclareMathOperator{\clip}{clip}
\DeclareMathOperator*{\argmax}{arg\,max}

\title{Beyond Search-Imitation:\\ Prior-Directed Exploration for Searchless Chess}
\author{Szymon Mi\l{}osz\textsuperscript{*} \quad Piotr Duch \quad Szymon Grabowski \\
    {\normalsize Institute of Applied Computer Science, Lodz University of Technology} \\
    {\normalsize \textsuperscript{*}Corresponding author:
     \texttt{szymon.milosz@dokt.p.lodz.pl}}}
\date{}

\begin{document}

\maketitle

\begin{abstract}
Searchless chess networks reach human master strength from a single forward pass by imitating a
stronger teacher: the strongest, Leela Chess Zero's (Lc0) released Chessformer, distills the visit
counts of an AlphaZero-style Monte Carlo Tree Search (MCTS). Imitating a search is a poor proxy for
playing without one, so we fine-tune for single-pass strength with self-play reinforcement learning
(RL). Its exploration is usually supplied by an entropy bonus---the reverse Kullback--Leibler (KL)
divergence to uniform. We replace it with a forward, mass-covering KL toward the network's own
MCTS prior---\emph{prior-directed exploration}---so exploration covers the moves the prior judges
promising, and pair it with an entropy-adaptive sampling temperature, set by the value head's
outcome uncertainty, that sharpens once a position is decided. In about two thousand steps it
raises puzzle accuracy from $93.9\%$ to $94.9\%$ on a $100{,}000$-puzzle suite and mate-in-four
accuracy from $77\%$ to $81\%$ while holding searchless strength at or slightly above the base.
Measuring tactical accuracy and playing strength together across a matched-compute sweep,
we find the two dissociate: accuracy gains fall in a
one-point band while ratings straddle the base, and a control fine-tuned on puzzles alone posts the study's largest tactical gains
while shedding roughly $260$ Elo---a better puzzle-solver is not thereby a stronger player.
Distribution-level measurements show what anchoring buys: without a regularizer self-play collapses
onto a single line of play, and the puzzles newly solved are the near misses whose winning
move the prior kept alive. The forward-KL prior tops the rating ladder, statistically tied with a
reverse-KL anchor that concentrates twice as hard and drops the hardest solutions the mass-covering
prior keeps in support.
\end{abstract}

\begin{quote}
\noindent\textbf{Code:} \url{https://github.com/szmilosz/prior-directed-exploration}
\end{quote}
\vspace{0.5em}

\section{Introduction}
\label{sec:intro}

\begin{figure}[t]
\centering
\includegraphics[width=0.72\linewidth]{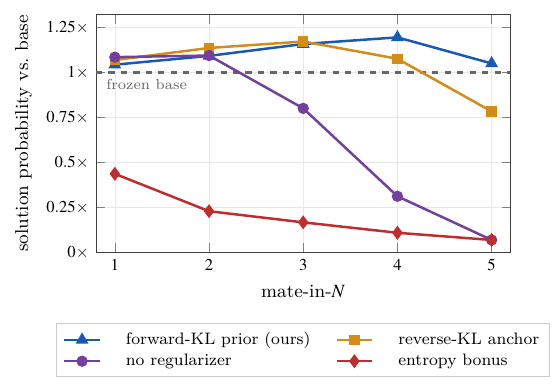}
\caption{What self-play fine-tuning does to the base's mate-finding, depth by depth: geometric-mean probability
$p^\star$ of sampling the full solution line from the model's output distribution, on the exact
mate-in-$N$ suites, relative to the frozen base
(dashed, $1\times$; ratios of the mate columns of Table~\ref{tab:nll};
Section~\ref{subsec:ablations}). Unregularized self-play sharpens shallow mates yet loses deep
ones ($0.07\times$ at depth five); the usual entropy bonus erodes every depth; anchoring
exploration to the network's own prior largely preserves them, and the mass-covering forward-KL
prior (ours) is the only arm above the base at every depth.}
\label{fig:hook}
\end{figure}

A recent line of work builds chess networks that play at up to human master strength\footnote{We use
``master'' conservatively and avoid ``grandmaster,'' which we think outruns the evidence. Strength
here is pool-dependent: \citet{ruoss2024amortized} rate the same searchless agent at $2895$ against
humans in the Lichess blitz pool but at $2299$ in an engine tournament, and the interpretability
study we build on titles our exact checkpoint ``grandmaster-level'' \citep{lin2026tracing}. Labels
of that kind rest on online blitz or engine-pool ratings: the only human-facing evidence in this
line is blitz play on Lichess, and the sole reported game record outside it is a five-game
exhibition against a $2100$-rated player. Online pools run well above the over-the-board scale at
equal strength, and none of these networks has been evaluated at classical time controls, where
such a claim would have to be settled.
Our claims are relative to the base network and do not turn on the absolute label.} while almost
eliminating test-time search: the move comes from a single forward pass of a policy
\citep{monroe2026chessformer}, a one-ply value lookup over the legal children
\citep{ruoss2024amortized}, or a fixed, shallow decoding or beam procedure standing in for a tree
search \citep{ye2025diffusearch,hamara2025planning}. Every one
of them is trained by imitation: a network is fit to a stronger but slower teacher and then
frozen. Our starting point is the strongest, the Chessformer architecture
\citep{monroe2026chessformer} as instantiated and trained by Leela Chess Zero (Lc0)
\citep{lc0}; it is obtained by supervised distillation of AlphaZero-style self-play data, fitting
its policy head to the search's MCTS visit-count distribution and its value head to game
outcomes. We write ``Chessformer'' for this released network from here on. Imitating a search is
mismatched with playing without one in two ways. The \emph{target} is mismatched, because the
visit-count distribution is only a noisy proxy for the move that maximizes win probability under
direct, searchless play. The \emph{state distribution} is mismatched, because the training games
were produced with search while the weaker searchless policy, left to its own devices, reaches a
different distribution of positions, the covariate shift familiar from imitation learning.
Self-play RL removes both at once: it optimizes a win/draw/loss reward for searchless strength on
data drawn from the searchless policy's own play. We use it to push Chessformer beyond its
search-imitation training.

Interpretability studies of Lc0's transformer chess networks make the target mismatch concrete:
these networks carry rich internal computation---with evidence even of learned look-ahead to
future optimal moves \citep{jenner2024lookahead}---yet on tactical puzzles with a single winning move,
that move is often computed in the intermediate layers only for the final layer to override it in
favor of a safer, positional alternative; a targeted intervention against this late-layer safety bias
has been shown to reverse the override for a majority of the suppressed wins \citep{sandmann2025iterative}
(observed on a smaller network of the same square-token lineage rather than on our exact checkpoint;
Appendix~\ref{app:basenet}).
Because searchless play commits to the $\argmax$, such an override costs strength outright, and we
conjecture that fine-tuning on game outcomes restores the move's rank---reaching through training the
correction their activation steering reaches at inference.

Reinforcement learning improves a policy only to the extent that it explores, and policy-gradient
methods sustain exploration with an entropy bonus that keeps the action distribution spread out
\citep{williams1992reinforce,williamspeng1991entropy,mnih2016a3c,schulman2017ppo,haarnoja2018sac}. The entropy bonus is
indifferent to \emph{which} moves are worth trying. Up to an additive constant it is the negative
KL divergence $-\KL{\pith}{\mathcal{U}}$ to the uniform distribution $\mathcal{U}$, so it pushes
probability onto all legal moves, including the many that no strong player would consider. With a
large branching factor this spends the exploration budget on obviously bad moves and gives a weak,
undirected signal; the same uninformed spreading, and its converse entropy collapse, are live
concerns in language-model RL \citep{cui2025entropy,wang2025scope}.

A far better reference is already available at no extra cost: the base network's own policy. The
network was trained to match the MCTS visit-count distributions of a self-play search, so a single
forward pass already returns a search-derived map of the moves worth examining---not the search
itself, but the network's distilled approximation of it. We use this policy $\pibase$ as the
exploration reference, replacing the entropy bonus with a forward, \emph{mass-covering} KL penalty
$\KL{\pibase}{\pith}$ toward it. Direction is what turns a divergence
into an exploration signal: the mass-covering KL diverges whenever the policy drops a move the
prior considers promising, forcing the policy to keep covering the prior's informed support,
whereas the mode-seeking $\KL{\pith}{\pibase}$, the anchoring penalty of reinforcement learning
from human feedback (RLHF) and from verifiable rewards (RLVR)
\citep{ouyang2022training,shao2024deepseekmath}, would let the policy collapse onto one move.
Using the mass-covering direction for exploration has appeared recently in language-model RL
\citep{deng2025rapo,li2025divergence}; we adapt it as a drop-in replacement for the entropy
bonus, toward a fixed search-derived prior, in searchless chess. We call the mechanism
\emph{prior-directed exploration}; Section~\ref{sec:related} places it among the alternatives.

Prior-directed exploration helps only when the reference is itself exploratory, spreading mass
over several promising moves rather than one, so that there is breadth worth covering.
Search-derived policies satisfy this by construction, which is why the MCTS prior plays a double
role here: a noisy \emph{pointwise} target for imitation, but an excellent \emph{exploratory}
reference for RL. The late-layer demotion described above \citep{sandmann2025iterative} illustrates
this split directly: the winning move the output passes over is not erased but re-ranked just below
the top, so it persists in the policy's distribution with substantial mass---failing as a pointwise
target yet remaining precisely the support a coverage-preserving penalty can keep while the reward
lifts it back (Section~\ref{subsec:recovery} measures this: the solution sits at rank two in $71\%$
of the base's failures). During fine-tuning the win/draw/loss reward re-sharpens the policy within the
prior's support toward what actually wins, while the forward KL keeps it from collapsing
(Figure~\ref{fig:hook} previews the outcome on exact mates).

\paragraph{Contributions.}
\begin{itemize}
\setlength{\itemsep}{1pt}
\item We cast strengthening a searchless network as moving \emph{beyond} search-imitation and
instantiate it as self-play RL that optimizes Chessformer's single-pass strength directly,
addressing the noisy-target and distribution-shift mismatches of imitation, in a single-step
self-play recipe: truncated importance sampling and a one-step temporal-difference (TD(0)) value
head with an exponential-moving-average (EMA) target
(Sections~\ref{subsec:instantiation}--\ref{sec:experiments}).
\item For exploration we replace the entropy bonus with a forward, mass-covering KL toward the
network's own MCTS visit-count prior, \emph{prior-directed exploration}, adapting a mechanism
recently used in language-model RL \citep{deng2025rapo,li2025divergence} to a fixed external
prior whose support is worth covering; we further argue (Appendix~\ref{app:prior}) that a
search-distilled prior is exploratory by construction (Section~\ref{subsec:structured}).
\item We introduce an \emph{entropy-adaptive sampling temperature} that sets each move's
temperature to the value head's win/draw/loss uncertainty, so self-play terminates once a
position is decided yet stays exploratory while it is live (Section~\ref{subsec:instantiation}).
\item Empirically, where the interpretability studies of these networks probe them almost entirely
through tactical puzzles, we show prior-directed fine-tuning improves \emph{both} tactics and
searchless strength---modestly on strength, in about two thousand steps.
We further \emph{quantify}, under matched compute, that tactical accuracy and playing strength
\emph{dissociate}, so the coupling rests on anchoring exploration to the prior rather than on
tactical optimization alone; and we show the tactical gains are \emph{retention-gated}:
fine-tuning promotes near misses the base still ranked second or third and recovers nothing whose
solution the base had driven below $\pibase\!\approx\!0.05$ (Section~\ref{sec:experiments}).
\item We define and measure two coverage-versus-collapse diagnostics---trajectory coverage
$\Omega^N_K$ and the \emph{solution probability} of sampling a ground-truth winning line in one
pass. Both split as the mechanism predicts: the unregularized policy collapses onto a single
opening line, and the mode-seeking anchor concentrates about twice as hard as the forward prior
and drops the hardest solutions, where the forward prior retains them
(Section~\ref{sec:experiments}).
\end{itemize}

\section{Method}
\label{sec:method}

\subsection{Background: policy gradients and entropy regularization}
\label{subsec:background}

We work with policy-gradient RL: a stochastic policy $\pith(a\mid s)$ is improved by ascending the
objective $\E_{(s,a)}[A(s,a)\log\pith(a\mid s)]$, with $A(s,a)$ an estimate of the advantage of the
sampled move. Unregularized, this
update is self-reinforcing. It keeps raising the probability of whatever currently looks best and
drives $\pith$ toward a deterministic policy that no longer samples alternatives, often before
the best move has been found. The failure is self-sealing: a policy-gradient update credits only the
moves it actually samples, so once a move's probability reaches zero its reward is never observed
again and no gradient can restore it---exploration must keep candidate moves sampled for their
rewards to remain discoverable. A regularizer that sustains exploration is therefore added; the
standard one is the entropy bonus $\lambda\,\E_s[\Ent(\pith(\cdot\mid s))]$ with $\lambda>0$. Up
to an additive constant it penalizes the KL divergence to the uniform distribution $\mathcal{U}_s$
over the legal moves $\mathcal{A}(s)$,
\begin{equation}
  \Ent(\pith(\cdot\mid s)) = \log|\mathcal{A}(s)| - \KL{\pith(\cdot\mid s)}{\mathcal{U}_s},
  \label{eq:entropy-kl}
\end{equation}
with the optimized policy as the \emph{first} argument. We follow the convention in which the
\emph{forward} (inclusive) KL places the fixed reference first and the \emph{reverse} (exclusive)
KL places the optimized policy first. Entropy regularization
is thus the reverse KL toward a single, fixed, and maximally uninformative reference. The choice
of direction, central to our method, is the subject of the next subsection.

\subsection{Prior-directed exploration via a forward-KL prior}
\label{subsec:structured}

The uniform reference encodes nothing about the task. When a strong policy $\pibase$ is on hand,
it describes far better where exploration should go, because it already places mass on the moves
worth trying. We replace the entropy bonus with a penalty that pulls $\pith$ toward $\pibase$ in
the forward, mass-covering direction,
\begin{equation}
  \mathcal{R}(\theta) = \E_s\!\left[\KL{\pibase(\cdot\mid s)}{\pith(\cdot\mid s)}\right]
  = \E_s\!\Big[\textstyle\sum_{a\in\mathcal{A}(s)}
  \pibase(a\mid s)\log\tfrac{\pibase(a\mid s)}{\pith(a\mid s)}\Big],
  \label{eq:kl}
\end{equation}
which, added with weight $\beta>0$ in place of the entropy bonus, gives the policy objective
\begin{equation}
  \mathcal{L}_\pi(\theta) = -\,\E_{(s,a)}\!\left[A(s,a)\log\pith(a\mid s)\right]
  + \beta\,\E_s\!\left[\KL{\pibase(\cdot\mid s)}{\pith(\cdot\mid s)}\right].
  \label{eq:policy}
\end{equation}

\begin{figure}[tbp]
\centering
\includegraphics[width=0.85\linewidth]{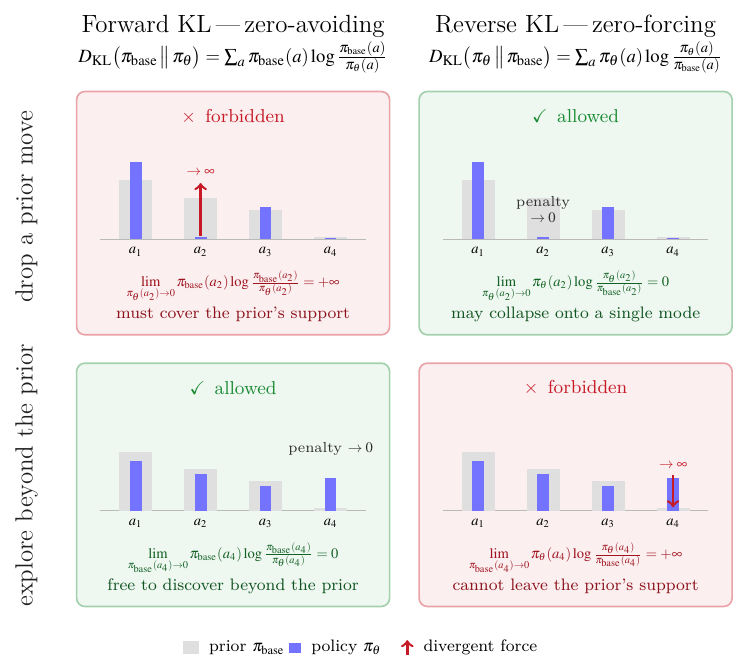}
\caption{The two asymmetries between the KL directions; probabilities are softmax outputs and
hence strictly positive, so the limits describe behavior as a probability becomes arbitrarily
small. \textbf{Rows:} the policy either drops a prior move ($\pith(a)\to0$ where $\pibase(a)$ is
large) or explores beyond the prior (mass where $\pibase(a)\to0$). \textbf{Columns:} the forward
KL of our penalty and the reverse KL of the standard anchor. The forbidden cells lie on the
diagonal: forward KL is zero-avoiding, blowing up when the policy drops a prior move yet free
where the prior is small; reverse KL is zero-forcing, the mirror image, permitting collapse onto
a single prior mode.}
\label{fig:mechanism}
\end{figure}

\paragraph{Why the forward direction.}
Direction is the crux, and Figure~\ref{fig:mechanism} lays out the asymmetry in full. Our penalty
$\KL{\pibase}{\pith}$ takes the forward, mass-covering direction---in the usual terminology
\emph{zero-avoiding}---so minimizing it forces the policy to keep
probability on every move the
prior deems promising, which is exactly what an exploration signal should do---it prevents collapse
to a near-deterministic strategy, and the moves it must cover are the prior's high-probability ones
rather than the whole action set. The reverse KL $\KL{\pith}{\pibase}$ is mode-seeking and
\emph{zero-forcing} \citep{minka2005divergence} instead---the anchoring penalty of KL-regularized policy optimization, from
behavior priors to RLHF and RLVR
\citep{galashov2019information,ouyang2022training,shao2024deepseekmath,tirumala2020behavior}.
That is the right property for the
anchor's job, which is to confine the policy to a trusted reference's support---in RLHF so that a
learned reward cannot be over-optimized by drifting off it (reward hacking)---but as an
exploration driver it would
license the very collapse we are trying to prevent: the two uses want opposite failure modes from
the same divergence. We are deliberate about the word
``exploration'': the forward KL sustains \emph{coverage} of the prior's promising moves and prevents
collapse onto one of them, but being zero where $\pibase(a)\to0$ it cannot drive discovery of strong
moves \emph{outside} the prior's support. It preserves prior-anchored breadth rather than seeking
novelty, unlike the intrinsic-motivation methods of Section~\ref{sec:related}.

\paragraph{Relation to entropy regularization, and the requirement on the prior.}
Entropy regularization is the \emph{reverse} KL to the uniform reference
(Equation~\eqref{eq:entropy-kl}), and even there the directions are not interchangeable: the
\emph{forward} KL to uniform $\KL{\mathcal{U}}{\pith}$ is the label-smoothing penalty
\citep{pereyra2017regularizing}, not the entropy bonus. Both share a minimizer against a uniform
reference, so either one merely spreads mass over all moves; prior-directed exploration keeps the
mass-covering direction and swaps that uninformative reference for the prior, turning uniform
spreading into coverage of the prior's promising moves. Because the penalty forces $\pith$ to
cover the support of $\pibase$, it is only as useful as that support: a near-deterministic expert
would pull the policy onto one move and reproduce collapse, so the method pairs naturally with
priors that are themselves products of exploration, such as MCTS visit-count policies.

Our prior meets that condition by construction, in two steps. First, the targets the search exports
are high-entropy exactly where exploration matters: the predictor-guided upper-confidence rule
(PUCT) keeps visiting every move with appreciable prior until its count catches up, so the exported distribution
$\pi_{\mathrm{mcts}}(a\mid s)\propto N(s,a)^{1/\tau}$ retains mass on several moves wherever more
than one is reasonable and sharpens only where the search itself is confident. Second, the released
network was fit to those targets by cross-entropy, which is the \emph{forward} KL
$\KL{\pi_{\mathrm{mcts}}}{\pibase}$ up to the target's own entropy---the same mass-covering
direction we use here---so a network that collapsed onto one move would pay a large loss at every
position where the search spread its visits, and the breadth transfers to the single-pass policy.
The prior we cover is therefore the product of a search, not a runtime search itself, and the
argument does not turn on the particular search variant; Appendix~\ref{app:prior} gives it in full.
How much coverage results under the full objective also depends on $\beta$ and the reward scale
\citep{gxchen2025mode}, which we set empirically.

A second enabling condition is computational: the forward KL needs the reference's full
distribution at each state, which is immediate for an enumerable categorical over the legal moves
but expensive for a language model over its vocabulary, one reason the reverse direction is the
default in large-scale post-training (Appendix~\ref{app:objectives}).

\FloatBarrier

\subsection{A single-step self-play recipe for searchless chess}
\label{subsec:instantiation}

We instantiate prior-directed exploration as self-play fine-tuning of Chessformer for searchless
play. The policy, its self-play rollouts, and the value bootstrap all use a single network
evaluation per position, with no tree search at training or test time, so the objective matches
searchless deployment. The reference $\pibase$ is the frozen Chessformer of
Section~\ref{sec:intro}; we initialize the online policy $\pith$ from it. The value head outputs a
win/draw/loss (WDL) distribution, summarized by the scalar score
$v_\theta(s)=p^{\mathrm{W}}_\theta(s)+\tfrac12 p^{\mathrm{D}}_\theta(s)\in[0,1]$. We estimate the
advantage of a move $a$ from $s$ to $s'=T(s,a)$ by a one-step TD(0) residual,
\begin{equation}
  A(s,a) = \big(1-v_{\bar\theta}(s')\big) - v_\theta(s),
  \label{eq:advantage}
\end{equation}
where $\bar\theta$ are exponential-moving-average (EMA) target weights and $1-v_{\bar\theta}(s')$
converts the successor score to the moving player's view, a flip licensed by the zero-sum,
symmetric self-play. Both $A$ and the importance weight introduced below are computed when the
transition is collected and enter the loss as constants, so no gradient reaches $v_\theta(s)$
through the policy term. The value loss regresses $v_\theta(s)$ toward the same bootstrapped
target that enters $A$, and the head starts near-consistent, so $\E[A]\approx0$ and mean-centering
would subtract almost nothing; we also avoid variance scaling, which would rescale the advantage's
natural win-probability units (Appendix~\ref{app:objectives}).
Data collection and optimization alternate synchronously: each batch is generated by the current
parameters and consumed by one optimizer step (gradients accumulated over micro-batches), with no
replay buffer and no additional epochs. The parameters that collect a batch therefore coincide with
the parameters it updates, the probability ratio of Proximal Policy Optimization (PPO) is
identically one, and PPO clipping \citep{schulman2017ppo} is inert:
the update reduces exactly to the advantage-weighted gradient of Equation~\eqref{eq:policy}
(derived in Appendix~\ref{app:objectives}). This is the familiar fact that one inner epoch turns PPO
into a baseline-corrected policy gradient; we present it as a simplification, not a stability
mechanism, since dropping the clip leaves the per-step change bounded only by the small layerwise
learning rates; the EMA target smooths the value estimates that feed $A$ but does not itself bound
the policy step.

\paragraph{Entropy-adaptive sampling temperature.}
The forward-KL prior shapes \emph{where} the policy explores; within a game we also control
\emph{how decisively} it plays. Sampling every move from $\pith$ leaves decided positions
needlessly stochastic and wastes rollouts that could begin fresh games; it also misdirects the
update, since converting a decided position is something the supervised base already does well,
and both the rollout budget and the learning signal are better spent on the contested positions
where the outcome is still in doubt. AlphaZero-style self-play
handles this with a ply-indexed temperature, sampling for the first plies and then playing
greedily \citep{silver2018alphazero}, but the move number is a coarse proxy for whether a position
is actually settled. We instead tie the temperature to the value head's own uncertainty, the
entropy of its WDL distribution,
\begin{equation}
  \tau(s) = \Ent(\wdl(s)) = -\!\!\sum_{o\in\{\mathrm{W},\mathrm{D},\mathrm{L}\}}\!\! p^o(s)\ln p^o(s)
  \in[0,\ln 3],
  \label{eq:temperature}
\end{equation}
and sample from the tempered behavior policy
\begin{equation}
  \mu(a\mid s) \propto \pith(a\mid s)^{1/\tau(s)}, \qquad a\in\mathcal{A}(s),
  \label{eq:behavior}
\end{equation}
with $\mu\to\argmax_a\pith(a\mid s)$ as $\tau\to0$.

\begin{figure}[tbp]
\centering
\includegraphics[width=0.66\linewidth]{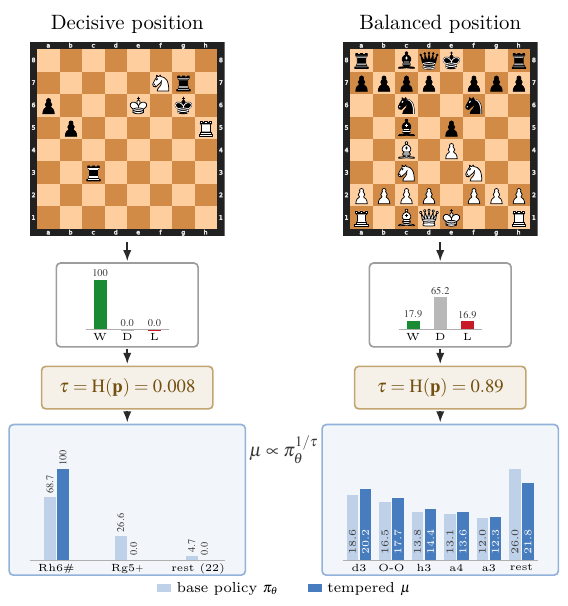}
\caption{Entropy-adaptive temperature, measured with the frozen base network: the value head's WDL
sets $\tau=\Ent(\wdl)$, and self-play samples from $\mu\propto\pith^{1/\tau}$. \emph{Left} (an
eight-piece rook-and-knight endgame, White to move, in which $\text{Rh6\#}$ is the only mate among
the $24$ legal moves): near-one-hot WDL, $\tau=0.008$, yet the base still keeps $26.6\%$ on
$\text{Rg5}+$, a check that does not mate, so sampling $\pith$ would miss the finish better than
one time in four; tempering discards it and commits to the mate ($68.7\%\to100\%$).
\emph{Right} (a balanced
opening): draw-heavy WDL, $\tau=0.89$, and $\mu$ barely sharpens the policy ($\mu\approx\pith$).
Light bars are the base policy $\pith$ and dark bars the tempered $\mu$ (left: the mate, the
strongest non-mating move, and the $22$ others pooled; right: the five most probable moves plus
the rest; per-panel scale, probabilities printed in \%).}
\label{fig:temperature}
\end{figure}
We use the raw entropy rather than normalizing
it: the maximum WDL entropy is $\ln 3\approx1.10$, so the unnormalized temperature spans
$[0,\ln 3]\approx[0,1.1]$, close enough to the unit interval to behave as a temperature; with $k$
outcomes one would divide by $\ln k$ to fix the range, which matters when $k$ is large. Because $\tau\le\ln 3$, the
exponent $1/\tau$ never falls below $0.91$: $\mu$ essentially never flattens $\pith$, it only
withdraws stochasticity from decided positions, and the spread it preserves elsewhere is the
spread the forward-KL term maintains. Decided positions have near-one-hot WDL, so $\tau\to0$ and
$\mu$ sharpens to finish the game quickly, reliable because the base value head is already strong
at conversion, while positions whose outcome is genuinely uncertain keep $\tau$ large and $\mu\approx\pith$, preserving
exploration where the learning signal is richest (Figure~\ref{fig:temperature}). The proxy is
outcome uncertainty, not move criticality, so $\mu$ also sharpens in the relatively rare position whose outcome
is settled yet whose move choice still matters; there it falls back on the base policy's top move,
dependable given the base's conversion strength but not infallible. Sampling from
$\mu$ rather than $\pith$ makes the batch off-policy, which we correct with a truncated importance
weight in the style of V-trace \citep{espeholt2018impala}; $\tau$ is computed from the EMA value
head for stability (Appendix~\ref{app:objectives}).

\begin{figure}[tbp]
\centering
\includegraphics[width=0.85\linewidth]{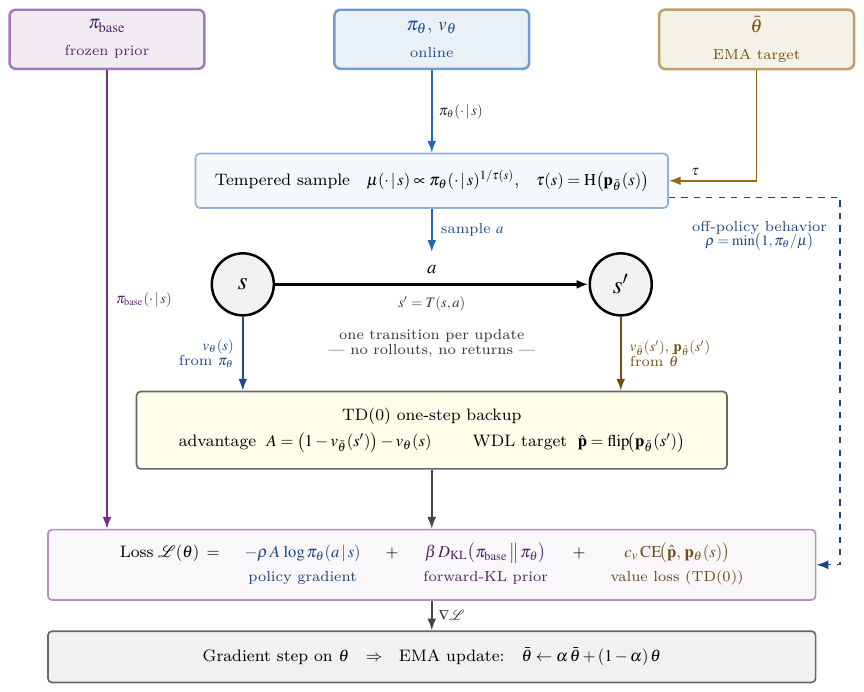}
\caption{Prior-directed self-play fine-tuning as a TD(0) one-step backup. Three networks: the
frozen prior $\pibase$ (the exploration reference), the online $\pith$, and an EMA copy
$\bar\theta$ supplying slow targets. At state $s$ the policy samples $a$ from
$\mu\propto\pith^{1/\tau(s)}$ with
$\tau(s)=\Ent(\wdl_{\bar\theta}(s))$ and steps to $s'$. Each update consumes this single
transition---no rollouts, no
returns: the online net evaluates $s$, the EMA bootstraps from $s'$, and one gradient step
minimizes the loss of Equation~\eqref{eq:total-loss}, after which the EMA tracks $\theta$. The dashed
wire is the off-policy gap the truncated weight $\rho$ corrects.}
\label{fig:algorithm}
\end{figure}

\newpage 
\paragraph{Overall objective.}
The value head is trained by the same one-step bootstrap: it regresses its WDL prediction at $s$
onto the point-of-view-flipped EMA prediction at $s'$ by cross-entropy $\mathcal{L}_V$
(Appendix~\ref{app:objectives}). The online network minimizes
\begin{equation}
  \mathcal{L}(\theta) =
  \underbrace{-\,\E_{(s,a)}\!\left[A(s,a)\log\pith(a\mid s)\right]}_{\text{policy gradient}}
  + \beta\,\E_s\!\left[\KL{\pibase}{\pith}\right]
  + c_v\,\mathcal{L}_V(\theta),
  \label{eq:total-loss}
\end{equation}
where the forward-KL term replaces the entropy bonus and $c_v>0$ balances policy and value.
Figure~\ref{fig:algorithm} diagrams this loop and Algorithm~\ref{alg:training} states it in
full; data generation, the truncated importance weight, and the single-step reduction are in
Appendices~\ref{app:selfplay} and~\ref{app:objectives}.

\begin{algorithm}[tbp]
\caption{Single-step self-play policy gradient with a forward-KL prior}
\label{alg:training}
\begin{algorithmic}[1]
\State Initialize online $\theta$ and EMA $\bar\theta$ from the reference $\pibase$
\State Launch $N$ parallel games, each seeded with $k\sim\mathrm{Uniform}\{0,\dots,60\}$ full moves from $\pibase$
\For{each training iteration}
  \State $\mathcal{B}\gets\emptyset$
  \For{each parallel game at state $s$}
    \State Evaluate $\pibase(\cdot\mid s)$; evaluate $\pith(\cdot\mid s)$ and $v_\theta(s)$
    \State $\tau(s)\gets\Ent(\wdl_{\bar\theta}(s))$;\quad $\mu(\cdot\mid s)\propto\pith(\cdot\mid s)^{1/\tau(s)}$
    \State Sample $a\sim\mu(\cdot\mid s)$; apply $s'\gets T(s,a)$; evaluate $v_{\bar\theta}(s')$
    \State $\rho\gets\min(1,\pith(a\mid s)/\mu(a\mid s))$;\quad $A\gets(1-v_{\bar\theta}(s'))-v_\theta(s)$
    \State $\hat\wdl(s)\gets\operatorname{flip}(\wdl_{\bar\theta}(s'))$ \;(or true result if $s'$ terminal)
    \State Add $(s,a,\rho,A,\hat\wdl(s),\pibase(\cdot\mid s))$ to $\mathcal{B}$
      \State If $s'$ is terminal, restart this game from the initial position
  \EndFor
  \State Take \emph{one} gradient step on $\theta$ minimizing $\mathcal{L}(\theta)$ of Equation~\eqref{eq:total-loss} over $\mathcal{B}$
  \State $\bar\theta\gets\alpha\,\bar\theta+(1-\alpha)\,\theta$
\EndFor
\end{algorithmic}
\end{algorithm}

\section{Experiments}
\label{sec:experiments}

We ask whether prior-directed exploration improves a \emph{searchless} engine over its base---in
tactical accuracy and in playing strength, without the regression that tactical optimization alone
can cause---holding the rest of the pipeline fixed. The experiments test four things: whether
self-play RL moves the supervised base beyond search-imitation in tactics and in strength; whether
tactical accuracy and playing strength move together or dissociate across exploration terms; how the
forward-KL prior compares with the entropy bonus, a reverse-KL anchor, and no regularizer at equal
compute; and how much each component of the recipe contributes.

\subsection{Setup and metrics}
\label{subsec:setup}

\paragraph{Setup and conditions.}
Starting from the same frozen Chessformer reference, the publicly released Lc0 checkpoint
\texttt{BT4-1024x15x32h-swa-6147500} (architecture in Appendix~\ref{app:basenet}), we fine-tune with the procedure of
Section~\ref{subsec:instantiation}, varying \emph{only} the exploration term and holding the
network, optimizer, self-play budget, and evaluation fixed. All play and evaluation are searchless.
Four fine-tuned conditions share that pipeline---(i) \emph{no regularizer}; (ii) the
\emph{entropy bonus}; (iii) a \emph{reverse-KL anchor} to $\pibase$; and (iv) the \emph{forward-KL
prior} (ours)---and all are measured against the \emph{frozen base}, the reference to beat. The four
are matched on regularization strength and compute. Each
arm's coefficient $\beta^\star$ is selected by overall accuracy on the $10{,}000$-puzzle suite described
under Metrics below---$10^{-2}$ for five of the six term--temperature arms, $10^{-4}$ for the sixth (the entropy
bonus without the adaptive temperature)---and every subsequent evaluation uses the selected model
(Table~\ref{tab:sweep}, Appendix~\ref{app:config}). Ratings carry $95\%$ confidence intervals (one run per cell;
Appendix~\ref{app:metrics}); the full training configuration is in Table~\ref{tab:hyperparameters} and Appendix~\ref{app:config}.

\paragraph{Ablation design.}
The runs cross two factors: the exploration term with its coefficient $\beta$, and the
entropy-adaptive sampling temperature against a fixed $\tau\equiv1$ baseline, so that the
temperature's contribution is separated from the exploration term's. For each
regularizer we sweep $\beta$ over the same five-point decade grid from $10^{-4}$ to $1$, so that
every arm is selected on identical footing (Appendix~\ref{app:config} lists the full matrix). Fixing the term and varying $\beta$ locates each
method's best operating point on that suite (Table~\ref{tab:sweep}); fixing $\beta$ and
varying the temperature isolates the adaptive temperature; the no-regularizer runs anchor both axes.

\paragraph{Metrics.}
Playing strength is reported two ways in Table~\ref{tab:main}: relative searchless Elo from a closed
round-robin, which resolves within-pool differences and reaches an absolute axis only through a
single external anchor on the base, and each model's direct record over $1{,}000$ games against the
base---its win--draw--loss (W--D--L) record, and the BayesElo likelihood of superiority (LOS) that it
is the stronger player. The protocol is in Appendix~\ref{app:metrics}.

Tactical accuracy is measured on three suites of Lichess puzzles. The first is the
$10{,}000$-puzzle suite released by \citet{ruoss2024amortized} and reported again in the
Chessformer evaluation \citep{monroe2026chessformer}; scoring on it keeps our base comparable with
the published figure for the same network (Appendix~\ref{app:basenet}), and we use it to select
each arm's coefficient (Table~\ref{tab:sweep}). Ten thousand puzzles are too few for the
comparisons this paper turns on, however: the $95\%$ Wilson interval at that size is
$\approx\!\pm0.45$ percentage points, while the fine-tuned arms separate by a few tenths of a
point. We therefore built two larger suites, sampling the public Lichess puzzle database the same
way: a uniformly sampled $100{,}000$-puzzle suite, which tightens the interval to
$\approx\!\pm0.15$ points, and a $20{,}000$-puzzle suite restricted to puzzles rated
$\geq\!2400$, which spends the sample where the base still fails and the arms still differ. Those ratings are Lichess's Glicko-2 estimates from human solve rates, so they order puzzles by human difficulty rather than by engine depth---a signal regular enough that predicting it from the board is itself a studied task \citep{milosz2024glickformer, milosz2025pretraining}. These
two carry every headline accuracy in the paper (Table~\ref{tab:main}).

Exploration versus collapse plays out over sequences, so to tie strength to mechanism we add two
sequence-level diagnostics, \emph{trajectory coverage} and the \emph{solution probability},
defined next.

\paragraph{Trajectory coverage $\Omega^N_K$.}
From a state $s$ we draw complete depth-$N$ trajectories from the policy's play
distribution \emph{without replacement}: a draw samples a continuation move by move, its trajectory
probability $\pith(\xi\mid s)=\prod_{t=1}^N\pith(a_t\mid s_{t-1})$ is recorded, and that mass is
removed from the tree and the remainder renormalized (implemented by subtracting the trajectory's
mass along its path), so the next draw samples among the lines not yet seen. The coverage after $K$
draws,
\begin{equation}
  \Omega^N_K(s) = \E\Big[\sum_{i=1}^{K} \pith(\xi_i\mid s)\Big],
  \label{eq:omega}
\end{equation}
with the expectation over the sampling and each $\pith(\xi_i\mid s)$ evaluated under the original
policy, is the total mass of the $K$ distinct lines drawn: a near-deterministic policy reaches
$\Omega^N_K\approx1$ within a handful of draws, while a policy that spreads mass over many promising
continuations needs thousands. Sweeping $K$ at fixed $N$ traces a coverage curve whose shape
distinguishes collapse from prior-directed spreading, the multi-step counterpart of the single-state
coverage and entropy diagnostics; sampled models are drawn through the same play distribution used in
self-play (the tempered $\mu$ for adaptive-temperature models), so the curve measures how quickly
self-play exhausts what the policy will actually play.

\paragraph{Solution probability, via the solution negative log-likelihood (NLL).}
Tactics accuracy scores whether the $\argmax$ move solves a puzzle, which is harsh on a stochastic policy
that ranks the solution highly without making it the single most probable move. Each puzzle specifies a
ground-truth line $\xi^\star=(a^\star_1,\dots,a^\star_M)$ visiting positions
$s^\star_0,\dots,s^\star_{M-1}$, with a set $W(s^\star_t)$ of winning moves at each position (usually
the single $a^\star_{t+1}$; where several moves win, all are credited). Per puzzle we compute the
solution NLL, the log-mass the policy keeps along the line,
\begin{equation}
  \ell(\xi^\star) = -\sum_{t=1}^{M} \log \!\!\sum_{a\in W(s^\star_{t-1})}\!\! \pith(a\mid s^\star_{t-1}),
  \label{eq:nll}
\end{equation}
in nats, and report the \emph{solution probability} $p^\star = e^{-\ell(\xi^\star)}$---the
probability of sampling the entire line $\xi^\star$ in one rollout (exactly
$\prod_{t}\pith(a^\star_t\mid s^\star_{t-1})$ when each position admits a single winning move).
Aggregation happens in the log domain: averaging $\ell$ over puzzles and exponentiating yields the
\emph{geometric}-mean solution probability, which we report unless stated otherwise. The geometric
mean is the right default for retention because the probability is a product along the line and
heavy losses should not be masked: a policy that drives one solution toward zero probability is
charged in full, where the \emph{arithmetic} mean can stay high on the strength of the
solutions it keeps (e.g., solution probabilities $0.9$, $0.899$, and $0.001$ across three puzzles give an arithmetic mean of $0.600$ but a geometric mean of only $0.093$)---a divergence between the two means is itself a collapse signal, which
Section~\ref{subsec:ablations} exploits. Paired model comparisons are likewise made in the log
domain (per-puzzle NLL differences, reported as probability ratios). We compute $\ell$ under the
untempered $\pith$ (EMA weights), so it measures the learned policy's intrinsic retention of the
solution rather than the tempered behavior policy's, and it rewards exactly the property
prior-directed exploration is meant to preserve: keeping the correct line in support.

\begin{table}[tbp]
\centering
\caption{Puzzle accuracy and searchless playing strength after fine-tuning from the same Chessformer
base, evaluated single-pass ($\argmax$ moves, EMA weights) at each arm's selected
$\beta^\star$ (Table~\ref{tab:sweep}); higher is better. \emph{Acc.}: the $100{,}000$-puzzle suite;
$\mathrm{Acc}_{\geq\!2400}$: the $20{,}000$-puzzle $\geq\!2400$ suite; both
multi-solution scored, with $95\%$ Wilson intervals. Accuracies are absolute; of the Elo column only
the base row is, anchored to the Chessformer paper's $2374\pm37$ \citep{monroe2026chessformer},
with every other row the difference $\Delta$ from it---so comparisons rest on the per-row
intervals, which the shared anchor leaves unchanged. The last row is the supervised puzzle-tuned
control of Section~\ref{subsec:dissociation}.}
\label{tab:main}
\footnotesize
\setlength{\tabcolsep}{3.9pt}
\begin{tabular}{llccccc}
\toprule
 & & & & & \multicolumn{2}{c}{vs.\ base} \\
\cmidrule(lr){6-7}
Exploration term & Self-play $\tau$ & $\mathrm{Acc}$ (\%) & $\mathrm{Acc}_{\geq\!2400}$ (\%) & Elo & W--D--L & LOS \\
\midrule
Frozen base \citep{monroe2026chessformer} & -- & $93.85 \pm 0.15$ & $60.01 \pm 0.68$ & $2374 \pm 37$ & -- & -- \\
\addlinespace
No regularizer          & adaptive          & $94.06 \pm 0.15$ & $60.50 \pm 0.68$ & $\Delta\,{+}9.8 \pm 8.2$ & 326--351--323 & $0.55$ \\
                        & $\tau\!\equiv\!1$ & $94.04 \pm 0.15$ & $60.80 \pm 0.68$ & $\Delta\,{-}0.5 \pm 8.3$ & 380--295--325 & $0.98$ \\
\addlinespace
Entropy bonus           & adaptive          & $94.90 \pm 0.14$ & $62.30 \pm 0.67$ & $\Delta\,{-}5.8 \pm 8.3$ & 331--323--346 & $0.28$ \\
                        & $\tau\!\equiv\!1$ & $94.00 \pm 0.15$ & $60.47 \pm 0.68$ & $\Delta\,{+}6.6 \pm 8.2$ & 344--335--321 & $0.81$ \\
\addlinespace
Reverse-KL anchor       & adaptive          & $94.83 \pm 0.14$ & $62.67 \pm 0.67$ & $\Delta\,{+}16.2 \pm 8.3$ & 388--314--298 & $>\!0.99$ \\
                        & $\tau\!\equiv\!1$ & $94.78 \pm 0.14$ & $62.53 \pm 0.67$ & $\Delta\,{+}9.6 \pm 8.2$ & 335--349--316 & $0.77$ \\
\addlinespace
Forward-KL prior (ours) & adaptive          & $94.87 \pm 0.14$ & $62.67 \pm 0.67$ & $\Delta\,{+}11.7 \pm 8.3$ & 385--267--348 & $0.91$ \\
                        & $\tau\!\equiv\!1$ & $94.72 \pm 0.14$ & $62.21 \pm 0.67$ & $\Delta\,{+}16.6 \pm 8.3$ & 361--318--321 & $0.94$ \\
\midrule
Puzzle-tuned control (supervised) & -- & $96.80 \pm 0.11$ & $67.95 \pm 0.65$ & $\Delta\,{-}258.0 \pm 9.1$ & 169--101--730 & $<\!0.01$ \\
\bottomrule
\end{tabular}
\end{table}

\begin{table}[tbp]
\centering
\caption{Training hyperparameters shared across all runs of Section~\ref{sec:experiments}. The
forward-KL weight $\beta$ is the only quantity that varies across conditions
(Section~\ref{subsec:ablations}).}
\label{tab:hyperparameters}
\small
\begin{tabular}{ll}
\toprule
Optimizer                              & AdamW, $\beta_1=0.9$, $\beta_2=0.98$ \\
Weight decay                           & $0$ \\
Schedule                               & 200-step linear warmup, then constant \\
Learning rate (backbone / final block) & $3{\times}10^{-6}$ / $1{\times}10^{-5}$ \\
Learning rate (policy head / value head)& $1{\times}10^{-5}$ / $3{\times}10^{-5}$ \\
Total steps                            & $2{,}000$ \\
Parallel games / batch size            & $1{,}024$ \\
Max plies per game                     & $512$ (drawn at cap) \\
Value-loss coefficient $c_v$           & $0.2$ \\
EMA decay $\alpha$                     & $0.995$ \\
Importance-sampling cap                & $1$ (V-trace truncation) \\
Forward-KL weight $\beta$ (headline)   & $10^{-2}$ (tactics sweep) \\
Gradient clipping / precision          & none / float32 \\
Dropout / stochastic depth             & off \\
\bottomrule
\end{tabular}
\end{table}

\subsection{Searchless playing strength}
\label{subsec:strength}

Because the gains are small, we read the ladder and the direct record together. Self-play moves
the base, but modestly. The forward-KL prior at fixed temperature tops the ladder at $+16.6\pm8.3$
Elo over the $2374$ base; head-to-head it outscores the base $0.52$ over $1{,}000$ games
($361$--$318$--$321$ W--D--L, likelihood of superiority $0.94$)---a real but slender edge, short of
decisive. The reverse-KL anchor is tied on the ladder ($+16.2\pm8.3$) and in fact wins its own match
against the base more clearly ($0.55$, LOS $>0.99$), so on strength the two prior-anchored directions
are indistinguishable and we do not claim the forward direction is the stronger \emph{player}. We read
the ladder cautiously for a second reason: a closed-pool rating and a direct record can
diverge---the unregularized fixed-temperature run rates near zero on the ladder yet outscores the base
$0.53$ head-to-head---because the rating reflects play against the whole pool, not the base alone.

Two
further cautions apply. The headline is the largest of many swept configurations, so part of its
margin is selection (winner's-curse) inflation the per-row interval does not capture. And the
exploration term's own contribution is modest and temperature-dependent: at fixed temperature the
forward-KL prior adds $\approx\!17$ Elo over no regularizer ($+16.6$ versus $-0.5$), but with the
adaptive temperature, which alone already reaches $+9.8$, it adds only $\approx\!2$ ($+11.7$ versus
$+9.8$)---prior and temperature are partly substitutes. What the strength results do establish is the
coarse ordering---anchoring exploration to the prior helps and clearly beats the fragile entropy
bonus, whose rating is acutely sensitive to its coefficient: without the adaptive temperature it
peaks at just $+6.6$ Elo (its selected $\beta=10^{-4}$) and falls to $-119$ by $\beta=10^{-2}$ and
$-576$ at $\beta=1$ (Table~\ref{tab:sweep} tracks the accompanying accuracy collapse)---while the forward-versus-reverse
question turns on move-distribution behavior the Elo cannot see, which
Sections~\ref{subsec:ablations} and~\ref{subsec:sampled} take up.

\begin{figure}[tbp]
\centering
\makebox[\linewidth][l]{\includegraphics[width=0.8966\linewidth]{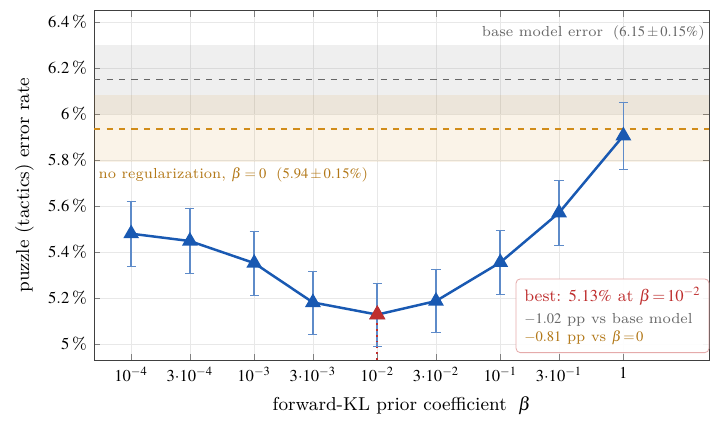}}
\caption{Forward-KL coefficient sweep of the headline adaptive-temperature arm, evaluated on
the $100{,}000$-puzzle suite: puzzle error of the EMA policy under temperature sampling versus the
forward-KL weight $\beta$ (lower is better; $\log\beta$ axis; error bars and shaded bands are
$95\%$ Wilson intervals). Dashed lines mark the frozen base and the unregularized $\beta=0$ run;
red marks $\beta=10^{-2}$, the coefficient selected on the $10{,}000$-puzzle suite. The four
half-decade points are additional resolution for this arm only.}
\label{fig:kl_sweep}
\end{figure}

\subsection{Tactical accuracy, retention, and coverage}
\label{subsec:ablations}

With the arms fixed by that selection, we turn from strength to the policies themselves: how
tactical accuracy responds to the coefficient, how much probability each policy retains on winning
lines, and how concentrated its play becomes.

\paragraph{Coefficient sweep.}
To trace the shape of the response rather than merely locate its optimum, we re-evaluate the
headline forward-KL-with-adaptive-temperature arm on the $100{,}000$-puzzle suite and add four
intermediate half-decade points, giving the nine-point curve of Figure~\ref{fig:kl_sweep}; these
extra points are for resolution only and play no part in selection. The larger suite's $95\%$
intervals of about $\pm0.14$ percentage points resolve differences the $10{,}000$-puzzle suite
could not. The
response to $\beta$ has a clean single minimum. The frozen base errs on $6.15\%$ of the puzzles and
the unregularized run ($\beta=0$) on $5.94\%$, so self-play already helps; adding the forward-KL
prior helps further, with the error falling to $5.13\%$ at $\beta=10^{-2}$, $1.02$ percentage
points below the base and $0.81$ below the unregularized run---both gaps several intervals wide.
The response is U-shaped in $\log\beta$ with its trough near
$\beta\in[3\cdot10^{-3},3\cdot10^{-2}]$, set between two limits the sweep approaches without
reaching: as $\beta\to0$ the objective reverts to the unregularized run ($5.94\%$), and as $\beta$
grows the penalty dominates the reward and pulls the policy back toward the base itself ($6.15\%$),
which it would recover exactly in the limit. Too small a weight therefore barely restrains collapse
and too large a weight over-regularizes: by $\beta=1$ the error has climbed to $5.91\%$,
surrendering the entire margin over the unregularized run on its way to the base. This slice confirms at ten times the scale the operating point
$\beta=10^{-2}$ selected on the smaller suite (Table~\ref{tab:sweep}). We do not over-read it:
it varies the forward KL and the adaptive
temperature jointly, on a tactics proxy, so it separates the forward-KL direction neither from the
adaptive temperature nor from value-head fine-tuning, and does not show that forward KL
beats the entropy bonus---a U-shaped response to a coefficient is generic to regularizers.

\paragraph{Accuracy by mate depth.}
Table~\ref{tab:mate} breaks searchless accuracy down by forced-mate
depth on the mate-in-$1$ to $5$
suite of \citet{lin2026tracing}. The base solves nearly every short mate and falls off with depth,
from $99.9\%$ at $N\!=\!1$ to $58.0\%$ at $N\!=\!5$; fine-tuning leaves the easy mates at ceiling and
concentrates its gains on the deeper mates ($N\!\ge\!3$), mirroring the $\geq\!2400$ puzzle column of
Table~\ref{tab:main}.

\begin{table}[tbp]
\centering
\caption{Searchless accuracy by forced-mate depth: $20{,}000$ Lichess positions per depth for
$N\!\le\!4$, from the suite of \citet{lin2026tracing}, and for $N\!=\!5$ all $4{,}465$ Lichess
puzzles whose forced mate is \emph{exactly} five moves (the Lichess mate-in-5 theme also contains
deeper mates), with the
network's true move history; a puzzle counts as solved only if the $\argmax$ at every position is a
mate-preserving move (any of several winning moves is credited). Models are the selected self-play
configurations of Table~\ref{tab:main}. All values are absolute accuracies with $95\%$
Wilson intervals; higher is better.}
\label{tab:mate}
\footnotesize
\setlength{\tabcolsep}{3pt}
\begin{tabular}{llccccc}
\toprule
 & & \multicolumn{5}{c}{Mate-in-$N$ accuracy (\%)} \\
\cmidrule(lr){3-7}
Exploration term & Self-play $\tau$ & $1$ & $2$ & $3$ & $4$ & $5$ \\
\midrule
Frozen base \citep{monroe2026chessformer} & -- & $99.94\pm0.04$ & $98.82\pm0.15$ & $92.58\pm0.36$ & $77.16\pm0.58$ & $58.01\pm1.45$ \\
\addlinespace
No regularizer          & adaptive          & $99.92\pm0.04$ & $98.77\pm0.15$ & $92.97\pm0.35$ & $79.23\pm0.56$ & $61.59\pm1.43$ \\
                        & $\tau\!\equiv\!1$ & $99.80\pm0.06$ & $98.90\pm0.14$ & $93.68\pm0.34$ & $80.73\pm0.55$ & $63.81\pm1.41$ \\
\addlinespace
Entropy bonus           & adaptive          & $99.95\pm0.03$ & $99.12\pm0.13$ & $94.13\pm0.33$ & $81.62\pm0.54$ & $64.28\pm1.40$ \\
                        & $\tau\!\equiv\!1$ & $99.91\pm0.04$ & $98.76\pm0.15$ & $93.08\pm0.35$ & $79.15\pm0.56$ & $61.55\pm1.43$ \\
\addlinespace
Reverse-KL anchor       & adaptive          & $99.95\pm0.03$ & $99.15\pm0.13$ & $94.21\pm0.32$ & $81.27\pm0.54$ & $63.31\pm1.41$ \\
                        & $\tau\!\equiv\!1$ & $99.93\pm0.04$ & $99.03\pm0.14$ & $94.03\pm0.33$ & $81.03\pm0.54$ & $62.96\pm1.42$ \\
\addlinespace
Forward-KL prior (ours) & adaptive          & $99.94\pm0.04$ & $99.18\pm0.13$ & $94.27\pm0.32$ & $81.38\pm0.54$ & $63.23\pm1.41$ \\
                        & $\tau\!\equiv\!1$ & $99.94\pm0.04$ & $99.02\pm0.14$ & $93.80\pm0.33$ & $81.06\pm0.54$ & $62.67\pm1.42$ \\
\bottomrule
\end{tabular}
\end{table}

\begin{figure}[tbp]
\centering
\includegraphics[width=\linewidth]{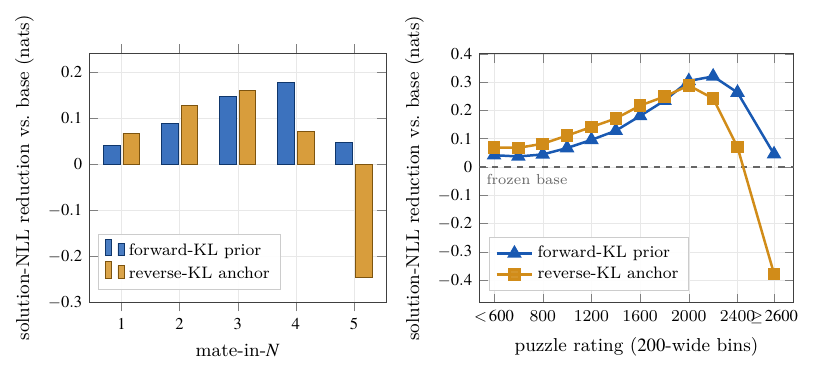}
\caption{The difficulty gradient of the two prior-anchored models: solution-NLL reduction over the
frozen base (base minus model, in nats---the log of the solution-probability ratio; above zero the
model retains solutions better than the base). \emph{Left:} by mate depth (the mate columns of
Table~\ref{tab:nll}). \emph{Right:} by $200$-Elo puzzle rating bin on the $100{,}000$-puzzle suite
(per-puzzle sequence NLL on the solution line); ratings below $600$ and above $2600$ are pooled
into single bins ($2{,}496$--$14{,}167$ puzzles per bin). The entropy model is not plotted, its losses lying far below both axes
(Table~\ref{tab:nll}).}
\label{fig:delta_bins}
\end{figure}

\paragraph{Solution retention.}
Argmax accuracy asks only whether the top move is right; a sharper question is what probability the
policy \emph{retains} on the full winning line---the chance of sampling the entire solution in one
pass, the quantity prior-directed exploration is meant to protect, and the one that keeps
improvement reachable during training (Section~\ref{subsec:background}). Table~\ref{tab:nll} reports
the geometric-mean $p^\star$ for the base and the fine-tuned conditions, and
Figure~\ref{fig:delta_bins} (left) plots the two prior-anchored models' gain over the base by mate
depth. The forward prior retains solutions better than the base on every
suite; the reverse anchor does too on short and mid-depth lines but falls \emph{below the base} on
the hardest sets ($0.178$ versus $0.199$ on the $\geq\!2400$ puzzles; $0.126$ versus $0.161$ at
mate-in-5). The two directions split exactly as the mechanism predicts. The mode-seeking reverse
anchor is sharper where lines are short, its retained probability $1$--$4\%$ higher (as a ratio) on
the full tactics suite and on mates in one to three, while the mass-covering forward prior wins
increasingly with difficulty and depth: its solution probability is $33\%$ higher on the
$\geq\!2400$ puzzles (paired ratio $1.33$) and $11\%$ and $34\%$
higher at mate-in-4 and 5. The $\geq\!2400$ subset
shows the collapse signature directly, in the two means of Table~\ref{tab:nll}: the reverse
anchor's \emph{arithmetic}-mean solution probability ($0.57$) tops the forward prior's ($0.53$) and
nearly doubles the base's ($0.30$), yet its geometric mean falls below even the base's---it seizes
some hard solutions outright while dropping others toward zero probability, where the forward prior
keeps nearly all of them in support. The unregularized run is the same failure at its limit: the
highest arithmetic mean in the study ($0.60$) against a geometric mean of $0.01$, a sixty-fold
gap. Binned by puzzle rating on the $100{,}000$-puzzle suite (Figure~\ref{fig:delta_bins},
right), the comparison crosses over near rating $2{,}000$: the reverse anchor is ahead on every
easier bin and the forward prior on every harder one, and on the $\geq\!2600$ bin the reverse
anchor's retention collapses well below the base's ($-0.38$ nats) while the forward prior's stays
near it.
Argmax accuracy follows the same gradient, with a turn at the top: both prior-anchored models gain
under a tenth of a point where the base already saturates below rating $1{,}000$, rising to
$+3.5$ points in the $2{,}400$--$2{,}600$ bin before falling back to $+1.8$ on the hardest puzzles,
where nearly half the base's failures stay out of reach.

The entropy bonus makes the third corner of the comparison, and it fails in the opposite way. Its
adaptive-temperature model---the strongest $\argmax$ puzzle-solver among the self-play runs
(Table~\ref{tab:main})---retains dramatically \emph{less} of every solution than the base does:
$0.222$ versus $0.660$ on the tactics suite, $0.103$ versus $0.199$ on its $\geq\!2400$ puzzles, and
$0.399$ versus $0.915$ even at mate-in-1, where its $\argmax$ accuracy is at ceiling (losses of
$0.83$--$2.71$ nats across the mate depths and $0.53$--$1.34$ nats across the rating bins). This is
not the reverse anchor's mode-seeking collapse but the uninformed spreading of
Section~\ref{subsec:background} made visible: the entropy bonus leaves the top move in place while
flattening the distribution around it, so the $\argmax$ stays right as the mass a training run would
need to sample drains away. The three regularizers thus separate cleanly at the distribution level
even where $\argmax$ metrics tie.

\begin{figure}[tbp]
\centering
\includegraphics[width=0.9\linewidth]{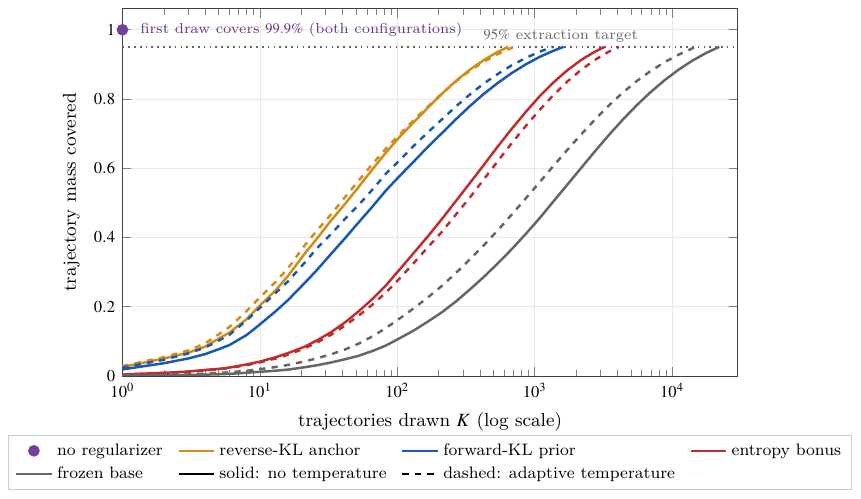}
\caption{Trajectory coverage $\Omega^{N=4}_K$ at the initial position: cumulative trajectory mass
covered by $K$ depth-$4$ continuations drawn without replacement from each model's play
distribution (mean over $100$ sampling seeds; all exploration terms at coefficient $0.01$, the
configurations of Table~\ref{tab:nll}). Solid: the no-temperature configurations (the frozen base
sampled raw); dashed: their adaptive-temperature counterparts (the base sampled through the
tempered $\mu$). Each curve ends where its extraction reaches the $95\%$ target (dotted); across
the $100$ seeds the standard deviation of the draws needed is at most $2.1\%$ of the mean.}
\label{fig:omega}
\end{figure}

\paragraph{Trajectory coverage.}
Retention asks about ground-truth lines; the trajectory coverage $\Omega^N_K$ asks how concentrated
the policy's own play is (Section~\ref{subsec:setup}). We draw depth-$4$ trajectories from the
initial position for every exploration term, in both temperature variants
(Figure~\ref{fig:omega}). Covering $95\%$ of the mass---out of $197{,}281$ legal
depth-$4$ continuations---takes about $22{,}000$ draws for the frozen base ($15{,}000$ through the
tempered $\mu$), $4{,}100$ for the entropy bonus, $1{,}400$ for the forward-KL prior, and $700$ for
the reverse-KL anchor ($3{,}200$, $1{,}600$, and $640$ for their no-temperature counterparts); the
unregularized run needs exactly \emph{one} in both variants, a single
opening line carrying $99.9\%$ of its mass---the self-sealing collapse of
Section~\ref{subsec:background} realized in full, and the sequence-level face of its retention
profile in Table~\ref{tab:nll}. Every regularizer prevents that collapse, and the survivors order
exactly as the mechanism predicts: the mode-seeking anchor concentrates about twice as hard as the
mass-covering prior, which in turn stays well inside the entropy bonus's near-uniform spread. All
fine-tuned policies are far more concentrated than the base---the expected effect of the reward
sharpening within the prior's support---so the diagnostic separates the exploration terms from one
another, not fine-tuning from its absence. The measurement is at a single position and depth;
Appendix~\ref{app:extra} repeats it at three further opening tabiyas of different character, where the
central ordering---reverse anchor tightest, forward prior wider, base widest by more than an order
of magnitude, the unregularized run collapsed---reproduces at each
(Figure~\ref{fig:omega-positions}).

\begin{table}[tbp]
\centering
\caption{Solution probability $p^\star$ (geometric mean over puzzles; higher is better), defined
in Section~\ref{subsec:setup}; mate suites use the network's true move history. Models are the
adaptive-temperature $\beta=10^{-2}$
configurations of Table~\ref{tab:main} (EMA weights); tactics columns as in Table~\ref{tab:main},
mate columns as in Table~\ref{tab:mate}. Best value per column in bold; fine-tuned values
are green above the frozen base, red below it (arithmetic means uncolored). For the $\geq\!2400$ suite the \emph{arithmetic} mean is
given in parentheses: the gap between the two means is the collapse signature read in
Section~\ref{subsec:ablations}. Figure~\ref{fig:delta_bins} (left) plots the mate columns.}
\label{tab:nll}
\small
\setlength{\tabcolsep}{5pt}
\begin{tabular}{lccccccc}
\toprule
 & \multicolumn{2}{c}{Tactics suite} & \multicolumn{5}{c}{Mate-in-$N$} \\
\cmidrule(lr){2-3}\cmidrule(lr){4-8}
Model & All & $\geq\!2400$ & $1$ & $2$ & $3$ & $4$ & $5$ \\
\midrule
Frozen base             & $0.660$ & $0.199$ ($0.302$) & $0.915$ & $0.819$ & $0.615$ & $0.344$ & $0.161$ \\
No regularizer          & \textcolor{tblloss}{$0.536$} & \textcolor{tblloss}{$0.010$} ($0.600$) & \textcolor{tblgain}{$\mathbf{0.993}$} & \textcolor{tblgain}{$0.896$} & \textcolor{tblloss}{$0.492$} & \textcolor{tblloss}{$0.107$} & \textcolor{tblloss}{$0.011$} \\
Entropy bonus           & \textcolor{tblloss}{$0.222$} & \textcolor{tblloss}{$0.103$} ($0.381$) & \textcolor{tblloss}{$0.399$} & \textcolor{tblloss}{$0.187$} & \textcolor{tblloss}{$0.102$} & \textcolor{tblloss}{$0.037$} & \textcolor{tblloss}{$0.011$} \\
Reverse-KL anchor       & \textcolor{tblgain}{$\mathbf{0.765}$} & \textcolor{tblloss}{$0.178$} ($0.569$) & \textcolor{tblgain}{$0.978$} & \textcolor{tblgain}{$\mathbf{0.930}$} & \textcolor{tblgain}{$\mathbf{0.721}$} & \textcolor{tblgain}{$0.370$} & \textcolor{tblloss}{$0.126$} \\
Forward-KL prior (ours) & \textcolor{tblgain}{$0.759$} & \textcolor{tblgain}{$\mathbf{0.236}$} ($0.526$) & \textcolor{tblgain}{$0.954$} & \textcolor{tblgain}{$0.894$} & \textcolor{tblgain}{$0.712$} & \textcolor{tblgain}{$\mathbf{0.411}$} & \textcolor{tblgain}{$\mathbf{0.169}$} \\
\bottomrule
\end{tabular}
\end{table}

\begin{figure}[tbp]
\centering
\includegraphics[width=\linewidth]{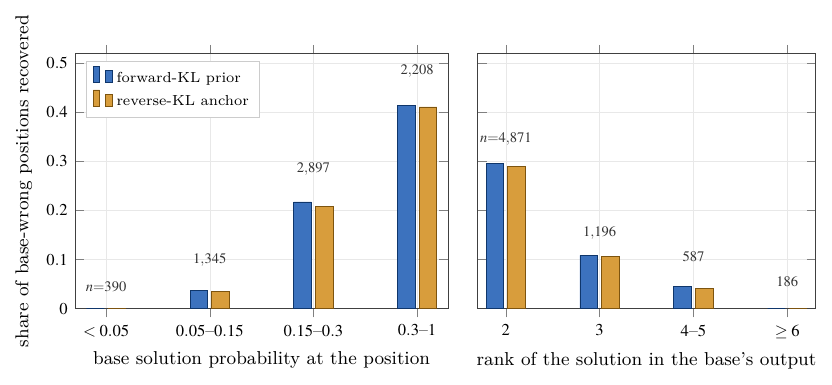}
\caption{Which failed positions fine-tuning recovers, for the two prior-anchored models of
Table~\ref{tab:nll} on the $100{,}000$-puzzle suite (scoring and totals in the text).
\textbf{Left:} share of base-wrong positions recovered, binned by the probability the base retains
on the solution at the position. \textbf{Right:} binned by the rank of the solution in the base's
output there; bin sizes annotated.}
\label{fig:recovery}
\end{figure}

\subsection{Policy shape: concentration, coverage, and sampled play}
\label{subsec:sampled}

The trajectory-coverage diagnostic of Section~\ref{subsec:ablations} measures how concentrated each
policy's play is; here we tie that concentration to strength and add three single-state
diagnostics of the policies themselves. We run a second closed round-robin over the
ten configurations of Figure~\ref{fig:omega-positions} in Appendix~\ref{app:extra}---the frozen base sampled raw and through the
tempered $\mu$, and each exploration term at both temperatures---in which every model \emph{samples}
its move rather than taking the $\argmax$, playing through the behavior policy it used in self-play
(Equation~\eqref{eq:behavior}; a $\tau\!\equiv\!1$ arm plays its raw $\pith$). Alongside the
resulting ratings, Table~\ref{tab:sampled} reports three diagnostics of the policy each model
actually plays: its move-distribution entropy, its two KL divergences to the raw base, and how much
of its mass stays on the base's top moves.

Under sampling, strength becomes a read-out of concentration. The ratings span $\approx\!670$ Elo,
more than twenty times the $\argmax$ ladder's $\approx\!30$ (Table~\ref{tab:main}), and they track
move entropy inversely: the more a policy concentrates, the better it plays when its moves
are sampled, because a diffuse policy squanders draws on its low-probability tail. This is a
diagnostic of the policies' shape, not a second measure of playing skill---the $\argmax$ ladder is the
strength result, and the sampled ordering (in which the reverse arm edges the forward by a margin
the $\argmax$ ladder does not show, tracking its slightly harder concentration in
Figure~\ref{fig:omega-positions}) merely re-scores how sharp each policy is. The adaptive temperature
is visible directly: tempering the base lifts it $+361$ Elo by sharpening its play ($\Ent$ from
$1.69$ to $0.39$), and every adaptive arm outranks its $\tau\!\equiv\!1$ twin.

The forward and reverse KL confirm the mechanism on played positions. Reading the $\tau\!\equiv\!1$
rows, whose eval policy is the raw learned $\pith$ and whose divergence to the base is therefore
comparable across arms, the forward-KL prior keeps $\KL{\pibase}{\pith}$ lowest ($0.22$), the
reverse anchor higher ($0.60$), and the entropy bonus higher still ($1.26$)---the mass-covering
penalty covers the base's mass two-to-six times more faithfully than the alternatives, the
zero-avoiding asymmetry of Figure~\ref{fig:mechanism} measured rather than argued. The tempered-$\mu$
rows carry a much larger forward KL (up to $11$) because tempering drives $\mu$ toward zero off the
top moves, which the base cannot cover---a property of the tempered behavior policy, not of the
learned $\pith$.

\begin{table}[tbp]
\centering
\caption{Sampled-play strength and single-state diagnostics for the ten configurations of
Figure~\ref{fig:omega-positions} in Appendix~\ref{app:extra} (the frozen base raw and tempered, and each exploration
term$\times$temperature arm; entropy is the $\beta\!=\!10^{-2}$ model). Sampling ``$\mu$'' is the
adaptive-temperature arm played through its tempered behavior policy, ``raw'' the $\tau\!\equiv\!1$
arm at temperature one. \emph{Elo}: sampled-play BayesElo from a closed round-robin (base anchored
$0$; $200$ games/pair, $45$ pairs; $95\%$ intervals). The remaining columns average, over the
positions each model reaches, its played policy $p$: entropy $\Ent(p)$ (nats), forward and reverse
KL to the raw base, and prior coverage $\mathrm{cov}_5$ (mass on the base's top-five moves).}
\label{tab:sampled}
\footnotesize
\setlength{\tabcolsep}{5.5pt}
\begin{tabular}{llrcccc}
\toprule
Configuration & Samp. & Elo & $\Ent(p)$ & $\KL{\pibase}{p}$ & $\KL{p}{\pibase}$ & $\mathrm{cov}_5$ \\
\midrule
Frozen base       & raw      & $0$          & $1.69$ & $0.00$  & $0.00$ & $0.81$ \\
                  & $\mu$    & $361\pm29$   & $0.39$ & $8.38$  & $0.81$ & $0.98$ \\
\addlinespace
No regularizer    & $\mu$    & $671\pm31$   & $0.06$ & $11.04$ & $1.03$ & $0.99$ \\
                  & raw      & $660\pm31$   & $0.22$ & $7.59$  & $0.89$ & $0.98$ \\
\addlinespace
Entropy bonus     & $\mu$    & $476\pm30$   & $0.74$ & $2.38$  & $0.59$ & $0.92$ \\
                  & raw      & $357\pm29$   & $1.44$ & $1.26$  & $0.26$ & $0.75$ \\
\addlinespace
Reverse-KL anchor & $\mu$    & $657\pm31$   & $0.28$ & $6.56$  & $0.80$ & $0.99$ \\
                  & raw      & $585\pm31$   & $1.52$ & $0.60$  & $0.18$ & $0.76$ \\
\addlinespace
Forward-KL prior  & $\mu$    & $602\pm31$   & $0.33$ & $6.42$  & $0.78$ & $0.99$ \\
                  & raw      & $501\pm30$   & $1.56$ & $0.22$  & $0.12$ & $0.77$ \\
\bottomrule
\end{tabular}
\end{table}

\subsection{What fine-tuning recovers}
\label{subsec:recovery}

The two prior-anchored models newly solve $1{,}405$ and $1{,}378$ puzzles the base
fails (and give back $383$ and $398$, for the net gains of Table~\ref{tab:main}), and the
retention machinery above predicts \emph{which}: improvement must be sampled before it can be
reinforced (Section~\ref{subsec:background}), so the failures fine-tuning can fix should be
exactly those whose solution the base still plays with non-negligible probability. That is what we
find (Figure~\ref{fig:recovery}). The policy is Markovian, so we score recovery at the level of
positions, under the multi-solution ground truth of Table~\ref{tab:main} (a move is correct if
it is an optimal mate or the marked move): on the $100{,}000$-puzzle suite the base plays a wrong
move at $6{,}840$ of its $231{,}739$ solver positions, and a position counts as recovered when the
fine-tuned model's move there is correct---the forward prior recovers $1{,}595$ of them, the
reverse anchor $1{,}560$. Binned by the probability the base retains on the solution at the
position, recovery is gated: neither model recovers any of the $390$ positions with
$\pi_{\mathrm{base}}(\text{solution})<0.05$, they recover about $4\%$ of the $1{,}345$ in
$[0.05,0.15)$, $21$--$22\%$ of the $2{,}897$ in $[0.15,0.3)$, and about $41\%$ of the $2{,}208$
above $0.3$ (geometric-mean retention of $0.30$/$0.31$ at recovered positions versus $0.17$ at
unrecovered). The rank view says the same thing more sharply: at $71\%$ of the base's wrong
positions the solution is ranked second, and these recover at $29\%$, falling to $11\%$ at
rank $3$ and about $5\%$ at ranks $4$--$5$; none of the $186$ positions with the solution at rank
$6$ or deeper is recovered by either model, and the deepest recovered solution sat at rank $5$.
The gap that had to be closed is correspondingly narrow: at recovered positions the base's top
move led the solution by a median of $6.5$ percentage points, against $24$ points at the
positions that stay wrong. Fine-tuning, under either
anchor, is \emph{near-miss promotion}: it reorders the top of the distribution the prior kept
alive, and it never resurrects a move the base had effectively discarded.
Recovery, where it happens, is decisive: the fine-tuned policies place a mean $36$/$42$
additional percentage points of mass on the recovered solutions (medians $35$/$43$), and the
puzzles gained are hard ones---mean rating $2{,}028$/$2{,}020$, far above the base's solved set
($1{,}429$) though below its average failure ($2{,}190$), as the retention gate predicts. The
give-backs are the same trade in reverse, still near misses (on $93\%$/$92\%$ the solution
remains at rank $2$ of the fine-tuned output) and shallow---the probability leaked off those
solutions has median $5.8$/$7.7$ points, against the $36$/$42$-point surges of the recoveries.
Give-back is gated from above as recovery is gated from below: of the more than $200{,}000$
positions where the base's top move leads the runner-up by over $0.6$, neither model gives back a
single one. Losing a puzzle is an $\argmax$ flip between two close moves; gaining one is a
commitment. The value
head moves with the policy: at recovered positions the base's win-minus-loss estimate has median
$0.02$---the value head barely sees the win it is failing to play---and fine-tuning lifts it to
$0.08$/$0.10$, while at unrecovered positions, where the base already reads the position more
clearly ($0.18$), it changes little ($0.17$/$0.19$).

\paragraph{Where the two anchors differ.}
The two anchors recover a large shared core and a small distinct fringe---$1{,}198$ puzzles both
fix, plus $207$ exclusive to the forward prior and $180$ to the reverse anchor---and the fringes
differ in kind. The forward prior's exclusive recoveries reach lower into the retained tail and
are the harder puzzles: median base solution probability $0.285$ and mean rating $2{,}139$,
against the reverse anchor's $0.297$ and $2{,}099$. This is the per-puzzle face of the retention
split of Table~\ref{tab:nll}: the mass-covering prior holds probability on hard, deep solutions
that the mode-seeking anchor lets go, and those are the wins only it can then promote.
Appendix~\ref{app:extra} walks through three exclusive recoveries of each anchor
(Figure~\ref{fig:examples}), annotating every candidate move with the base's own value-head
verdict and a Stockfish oracle; in five of the six positions the base's value head already
separates the winning move from the played one, so the first time self-play samples the retained
move, the advantage signal is immediate.

\subsection{Tactical accuracy versus playing strength}
\label{subsec:dissociation}

That any single evaluation figure is an imperfect proxy for playing strength is itself
known---a searchless transformer's rating swings some $600$ points between the human and engine
pools it is measured in \citep{ruoss2024amortized}, and the imitation-versus-strength split is the
premise of \citet{mcilroy2020maia}---but our matched-compute sweep lets us \emph{quantify} how wide
the tactics-versus-strength gap is here, and it is wide. Across the
eight fine-tuned configurations of Table~\ref{tab:main} the two axes move independently: every one
that improves overall accuracy does so within a narrow band ($+0.15$ to $+1.05$ percentage points)
while ratings range from $-5.8$ to $+16.6$ Elo and straddle the base. The clearest case
among the self-play runs is the entropy bonus with adaptive temperature, which posts the
\emph{highest} overall puzzle accuracy of any of them ($94.90\%$, $+1.05$ percentage points) yet earns \emph{no}
playing strength: its rating
($-5.8$ Elo) is statistically indistinguishable from the base, its interval spanning zero and its
head-to-head score $0.49$. Pushed harder the split turns to caricature: an entropy coefficient of $1$
still answers $83.1\%$ of the $10{,}000$-puzzle suite correctly (Table~\ref{tab:sweep}) while
losing $933$ of its $1{,}000$ games to the base.

Starker still is a control that optimizes puzzles \emph{directly}: fine-tuning the same base for the
same $2{,}000$ steps by supervised learning on the puzzle solutions rather than by self-play RL
(Table~\ref{tab:main}, last row). Its training pool is drawn from the Lichess puzzle database with
both evaluation suites removed, so it is scored on positions it has never seen. It
posts by far the largest tactical gains in the study---overall accuracy
$93.85\!\to\!96.80\%$
($+2.95$ percentage points) and on the $\geq\!2400$ suite $60.01\!\to\!67.95\%$ ($+7.94$ percentage points), around three times the
largest gain from any RL condition---yet its searchless rating \emph{collapses} to $-258.0\pm9.1$ Elo,
roughly a one-in-five score against the base. The cause is the state-distribution mismatch of
Section~\ref{sec:intro}, here induced on purpose: supervising on a fixed external set of positions
rather than on states from the policy's own play sharpens move-by-move tactics while whole-game
strength falls apart. The lesson is not that puzzle accuracy is a bad measure---it is the direct
measure of the tactical skill we target, and the dense, cheap signal our coefficient sweeps rely
on---but that it measures a \emph{different capability} than playing strength: most fine-tuning
recipes lift tactics, while strength is what separates a prior-anchored run that gains a little from
an entropy run that does not gain at all and a puzzle-tuned control that collapses. An evaluation
must therefore carry both axes separately. The forward- and reverse-KL priors are the configurations
that hold both up at once, raising tactical accuracy while keeping strength at or above the base,
and that coupling---not the puzzle figure on its own---is the result that bears on searchless play.

\FloatBarrier

\section{Related Work}
\label{sec:related}

\paragraph{Searchless chess.}
Classical neural engines couple a network with a large test-time search; a recent line almost
removes it, its strongest members playing at human master level. \citet{ruoss2024amortized} distill Stockfish
action-values into a transformer used as a depth-one move selector, one forward pass per legal
child with no recursion; DiffuSearch folds an implicit search into a discrete-diffusion policy
\citep{ye2025diffusearch}; \citet{hamara2025planning} navigate a contrastively learned value
embedding; and the Chessformer architecture \citep{monroe2026chessformer} as trained by Lc0
\citep{lc0}, our base and, in engine-versus-engine play, the strongest of these, selects each move from a single forward pass of
its policy head. A related thread interprets how such networks compute without explicit search
\citep{jenner2024lookahead,sandmann2025iterative,lin2026tracing}, and a separate imitation line fits \emph{human} move
distributions for human-like play rather than maximal strength \citep{mcilroy2020maia}; a further human-centric line predicts how \emph{hard} a position is for people, regressing Lichess's Glicko-2 puzzle ratings from the board \citep{milosz2024glickformer, milosz2025pretraining}. The playing systems in this line are
obtained by imitation and then frozen.

\paragraph{RL for games and pretrained policies.}
AlphaZero learns strong policies by self-play guided by MCTS \citep{silver2018alphazero}, extended by
MuZero to planning with a learned model \citep{schrittwieser2020muzero}, and Lc0 trains openly
available networks of this kind \citep{lc0,monroe2026chessformer}. That self-play loop
\emph{produces} the search-trained network we start from; it does not optimize the network's own
searchless play. Separately, RL adapts strong \emph{pretrained} policies, as in RLHF and RLVR, which
fine-tune language models against a reward with a KL anchor to the base
\citep{ouyang2022training,shao2024deepseekmath}. We join the two: we take an externally strong
searchless network and fine-tune it by self-play RL whose objective is searchless strength itself,
neither proposing a new way to obtain the network nor merely anchoring it to its initialization.

\paragraph{Exploration regularizers, and the three roles of the KL.}
An entropy bonus is the standard way to maintain exploration and avoid deterministic collapse
\citep{williamspeng1991entropy,mnih2016a3c,schulman2017ppo}, promoted to part of the objective in
maximum-entropy RL \citep{haarnoja2018sac}. Intrinsic-reward methods drive exploration with novelty
signals, count-based bonuses \citep{bellemare2016unifying}, prediction-error curiosity
\citep{pathak2017curiosity}, and random network distillation \citep{burda2019rnd}, which modify the
reward rather than the action-distribution regularizer and are orthogonal to what we study. A closer
neighbor is GFlowNets' mass-covering objective of sampling in proportion to reward
\citep{bengio2021gflownet}, which pursues the same anti-collapse, multi-modal coverage we seek but
enforces it by flow-matching rather than a divergence penalty. Entropy collapse is now a central
concern in language-model post-training \citep{cui2025entropy,wang2025scope}.
\citet{gxchen2025mode} qualify the direction argument we rely on: under KL-regularized RL the
direction fixes only the \emph{family} of optimal distributions, with mode coverage governed mainly
by the regularization strength and the scale of rewards relative to reference probabilities---which
is why we select $\beta$ empirically and rest the forward-versus-reverse case on the measured
coverage of Section~\ref{subsec:ablations} rather than on direction alone. The KL itself
plays three distinct roles in policy optimization: it constrains each update to the \emph{previous}
policy as a trust region \citep{kakade2002natural,schulman2015trust,schulman2017ppo}; it regularizes
toward a \emph{fixed} reference or learned default, generalizing maximum-entropy RL
\citep{galashov2019information,tirumala2020behavior}, underpinning control-as-inference
\citep{levine2018reinforcement} and maximum a~posteriori policy optimisation (MPO)
\citep{abdolmaleki2018maximum}, and anchoring fine-tuned
models in RLHF and RLVR \citep{ouyang2022training,shao2024deepseekmath}; and, in our use, it drives
exploration. Closest to our setting, \citet{ota2026revisiting} combine the first two roles for
two-player zero-sum games---a reverse KL to the current iterate together with an entropy
bonus---prove local convergence, and evaluate on five board games including a reduced chess; their
agents learn from scratch by self-play, where we adapt an externally strong released network, and
their entropy term points at the uniform distribution where ours points at that network. The
fixed-reference role is conventionally taken in the policy-first, mode-seeking direction
$\KL{\pith}{\pi_0}$, to keep the policy \emph{near} its reference; the trust-region role is taken in
either direction, reference-first in trust region policy optimization (TRPO) and PPO's
KL-penalty variant and policy-first in
\citet{ota2026revisiting}. The exploration regularizers
above all point at the uniform distribution. We keep the reference-policy regularizer, reverse its
direction, and swap its reference for a fixed, externally strong, search-derived prior.

\paragraph{Forward-KL regularization, and the closest work.}
Forward, mass-covering KL toward a reference is well established, but toward ends other than
exploration. Label smoothing adds a forward KL to the uniform distribution
\citep{szegedy2016rethinking,pereyra2017regularizing}; toward a strong reference, $f$-DPO generalizes
direct preference optimization (DPO) \citep{rafailov2023direct} to $f$-divergences including the
forward KL \citep{wang2024beyond}, and knowledge distillation matches a teacher under the forward KL
\citep{hinton2015distilling}, which on-policy distillation generalizes to a family of divergences
scored on student-generated data \citep{agarwal2024gkd}. In all of these the divergence
serves regularization, alignment, or distillation, not exploration. Closest to us, two recent
language-model works turn the forward direction toward exploration: diversity-preserving hybrid RL
(DPH-RL) anchors mass-covering
$f$-divergences to the model's \emph{own initial} policy as a rehearsal term against diversity
collapse \citep{li2025divergence}, and rewards-aware policy optimization (RAPO) replaces the reverse-KL anchor with a forward KL to
explore \emph{beyond} the base model's support, adding a reward-aware reweighting of the reference
to steer exploration \emph{within} it \citep{deng2025rapo}. We differ on three points. We replace the \emph{entropy bonus}, casting
prior-directed exploration as a generalization of entropy regularization rather than a variant of
the trust-region anchor. Our reference is a fixed, externally strong, search-derived prior used
\emph{as is} to cover its already-informed support, not the model's own initial policy or a
reference reweighted to redirect exploration. And our domain is searchless chess, where the target is
playing strength rather than reasoning accuracy.

\section{Conclusion}
\label{sec:conclusion}

Searchless chess networks are trained to imitate a search and then deployed without one. We argued
that self-play RL is the natural way to close this gap, optimizing single-pass strength directly,
and that the exploration it needs is better served by the network's own search-derived prior than by
an entropy bonus: replacing the bonus with a forward, mass-covering KL toward the MCTS visit-count
prior drives the policy to cover exactly the moves the prior judges promising, avoiding both
uninformed spreading and mode collapse, while the reward sharpens within that support. In about two
thousand steps, far below the search that produced the network, this delivers a clear tactical
improvement---higher puzzle accuracy and deeper forced mates---while searchless playing strength
holds at or just above the base, the forward-KL prior topping the rating ladder with the reverse-KL
anchor statistically tied.

The finding we underline constrains how such fine-tuning should be read: tactical accuracy and
playing strength dissociate, so a method that lifts puzzle accuracy is not thereby a stronger
player, and the worth of anchoring exploration to the prior is that it improves tactics without
spending strength. The distribution-level measurements separate the two anchoring directions as the
mechanism predicts---the mode-seeking anchor concentrates about twice as hard and drops the hardest
solutions, where the mass-covering prior keeps them, and without any regularizer self-play collapses
onto a single line of play---while the per-puzzle recovery analysis says what fine-tuning can and
cannot fix: it promotes the near misses the prior kept alive and never resurrects a move the base
had discarded, which is why keeping candidates sampled is the exploration term's whole job.

Chess is a
clean testbed because its search-derived priors are unusually strong. The recipe should extend to
settings that share its two enabling conditions, an \emph{enumerable} per-state reference and an
\emph{exploratory} prior worth covering, most naturally other games and combinatorial planning with a
search- or ensemble-derived prior. The standard language-model setting is not excluded in
principle but meets the enumerability condition only at a cost: the forward KL needs the reference's
full next-token distribution, the $O(|V|)$ teacher pass that on-policy distillation already pays
\citep{agarwal2024gkd} but that is dear to add on top of RL exploration (Section~\ref{subsec:structured})---so
chess serves here to exhibit the mechanism where the prior is cheap and strong, not to rule language out.

\paragraph{Limitations.}
Our claims are scoped to one domain and one base network. The forward-versus-reverse evidence supports
the mechanism on both retention and coverage, but the two directions remain tied on strength,
and the strength gains are small in absolute terms---a reminder that two thousand steps of
fine-tuning adjust the base network rather than remake it.

The method carries two structural requirements. The forward KL is computed exactly only for a categorical distribution over a modest action set; large action spaces require estimation, the cost
that makes the reverse direction the default there. And it helps only when the prior's support is
worth covering: a weak or biased prior would be faithfully covered, so annealing $\beta$ or combining
references is a natural extension. Neither
evaluation axis is a reliable arbiter alone either---self-play Elo can flatter a pool with shared blind
spots, and tactical accuracy can rise even for a weaker player---so we claim improvements only where both
agree.

Two risks are
specific to the self-play recipe. Because terminated games restart from the standard initial position,
a low-temperature, prior-anchored policy could narrow opening diversity over training; persistent root
randomization, as in AlphaZero, is the natural safeguard. Figure~\ref{fig:omega} quantifies the
narrowing after training---real for every run but far from collapse under any regularizer---though
it measures the policy's play distribution, not the opening diversity of the games actually
generated. And the
frozen prior is least reliable on exactly the off-distribution states self-play reaches, where forcing
coverage of its support is least justified---the covariate shift reappearing one level up---which
annealing $\beta$ as the policy drifts would mitigate.

\setlength{\bibsep}{6pt plus 1pt minus 1pt}
\bibliographystyle{plainnat}
\bibliography{references}

\appendix

\section{Notation, MDP, and networks}
\label{app:notation}

We model chess as an episodic, two-player, zero-sum Markov decision process. A state $s$ comprises
the board, the side to move, and the auxiliary rule information (castling rights, en-passant target,
repetition history). The agent selects a legal move $a\in\mathcal{A}(s)$; dynamics are deterministic,
$s'=T(s,a)$. Episodes end by the laws of chess with an outcome in
$\{\text{win},\text{draw},\text{loss}\}$ from the mover's view; intermediate rewards are zero and the
discount is $\gamma=1$, so an episode's return is its terminal score.

The network has a shared trunk $\theta$ and two heads. The policy head defines $\pith(\cdot\mid s)$
over $\mathcal{A}(s)$. The value head outputs a categorical WDL distribution $\wdl_\theta(s)=
(p^{\mathrm{W}}_\theta(s),p^{\mathrm{D}}_\theta(s),p^{\mathrm{L}}_\theta(s))$ from the mover's view,
giving the scalar score
\begin{equation}
  v_\theta(s) = p^{\mathrm{W}}_\theta(s) + \tfrac12\,p^{\mathrm{D}}_\theta(s) \in [0,1]
  \label{eq:scalar-value}
\end{equation}
(note that we assume the standard 1-0.5-0 scoring, though alternative rules rewarding decisive games, such as the 3-1-0 football-style system historically applied in events like the Bilbao Chess Masters, are sometimes used to discourage quick draws).

After a move the opponent is to move, so the mover's score is $1-v(s')$, with WDL vector obtained by
swapping the win and loss components, $\operatorname{flip}(\wdl)=(p^{\mathrm{L}},p^{\mathrm{D}},p^{\mathrm{W}})$.
We keep three copies of the network. The \emph{online} model $\pith$ is the only one updated by
gradient descent and is initialized from the reference. The \emph{reference} $\pibase$, the frozen
Chessformer, is the target of the forward-KL penalty. An \emph{EMA} model $\bar\theta$ tracks the
online weights,
\begin{equation}
  \bar\theta \leftarrow \alpha\,\bar\theta + (1-\alpha)\,\theta, \qquad \alpha\in(0,1),
  \label{eq:ema}
\end{equation}
serving as a slowly-moving target as in deep Q-learning \citep{mnih2015human}; decoupling bootstrap
targets from the rapidly changing online parameters is what stabilizes value learning.

\section{Base network: Chessformer}
\label{app:basenet}

The network we fine-tune is the Chessformer architecture \citep{monroe2026chessformer} in the
instantiation trained and released by Leela Chess Zero \citep{lc0}; every run uses the checkpoint
\texttt{BT4-1024x15x32h-swa-6147500} named in Section~\ref{sec:experiments}, a $15$-layer
encoder-only transformer with $1024$-dimensional embeddings, $32$ attention heads, and a
multi-layer-perceptron (MLP) width
of $1536$. It matches the $191$M-parameter \emph{Leela-CF} model of \citet{monroe2026chessformer} in
every dimension that work reports, and in accuracy on the $10{,}000$-puzzle suite the two share
($93.5\%$; Table~\ref{tab:sweep}), which is what licenses us to read that work's rating as this
checkpoint's (Appendix~\ref{app:metrics}). A position is
presented as $64$ tokens, one per board square, each a depth-$112$ feature vector encoding the
current and past $7$ board positions (piece occupancies, repetition flags, castling rights, side to
move, and the halfmove clock). Self-attention is augmented not by a generic positional encoding but
by \emph{Geometric Attention Bias} (GAB), the architecture's dynamic position encoding: from a
compressed representation of the board it generates per-head additive biases over the $64\times64$
square pairs, added to the dot-product logits before the softmax, so that attention adapts to the
geometry of the position \citep{monroe2026chessformer}; the Lc0 codebase calls the same module
\emph{smolgen}. The policy head is attention-based---source
and destination square representations are multiplied along the embedding axis to form a
$64\times64$ matrix of move logits, which restricted to the legal moves gives
$\pith(\cdot\mid s)$---and the value head mean-pools the trunk to predict the win/draw/loss
categorical $\wdl_\theta(s)$, summarized by the scalar score $v_\theta(s)$ of
Appendix~\ref{app:notation}.

The interpretability results we draw on in Section~\ref{sec:intro} are obtained on a different,
smaller Lc0 transformer of the same square-token lineage---the $768$-dimensional, $24$-head
\texttt{T82-768x15x24h} network ($\approx\!109$M parameters)
\citep{sandmann2025iterative}---not on the checkpoint we fine-tune. It nonetheless shares the
encoder-only, $64$-square-token form, a $15$-layer residual stream, and the generated
attention-bias module described above with our network, so the
layerwise behavior it reports---a strong move identified in intermediate layers and then demoted by
the output---is most naturally read as a property of these deep square-token policies rather than of
the exact weights we adapt.

The released network is obtained by \emph{supervised distillation}, not by an online
reinforcement-learning loop: its policy head is fit by cross-entropy to the Monte Carlo Tree Search
visit-count distributions of AlphaZero-style self-play, and its value head to the win/draw/loss game
outcomes \citep{silver2018alphazero}. The data originate from self-play \emph{with} search, but the
network itself only imitates the recorded search statistics, and it is not distilled from a classical
engine's action values---the distinction from the line of \citet{ruoss2024amortized}. This
supervised, search-derived origin is what makes the single-pass policy both the noisy pointwise
imitation target we move beyond and, as Appendix~\ref{app:prior} argues, an exploratory prior whose
support is worth covering.

\section{Why the base network's prior is exploratory}
\label{app:prior}

Prior-directed exploration helps only if the reference $\pibase$ spreads probability over several
promising moves rather than concentrating on one (Section~\ref{subsec:structured}).
This appendix explains why the base network meets that condition by construction. The
point to keep in view is that $\pibase$ is not an MCTS search performed at play time but the trained
network's single-pass policy, fit during pretraining to imitate the visit-count distributions of
such a search. The argument has two steps: those visit-count distributions are exploratory by
construction, and the supervised objective that produced the network is mass-covering, so the network
inherits their breadth.

\paragraph{Step 1: MCTS visit counts are exploratory by construction.}
AlphaZero-style training generates its data with MCTS \citep{silver2018alphazero}. From a position
$s$ the search repeatedly descends the tree, at each node taking the action that maximizes a PUCT
score that adds an exploration bonus to the running value estimate,
\[
  \argmax_{a}\; Q(s,a) + c_{\mathrm{puct}}\,P(s,a)\,\frac{\sqrt{\textstyle\sum_b N(s,b)}}{1+N(s,a)},
\]
so a move with appreciable prior $P(s,a)$ keeps being tried even after another currently looks best,
until its visit count $N(s,a)$ catches up. The policy the search exports is not its single best move
but the (optionally tempered) visit-count distribution
$\pi_{\mathrm{mcts}}(a\mid s)\propto N(s,a)^{1/\tau}$. Because the bonus spreads visits across every
move with non-negligible prior and value, this distribution keeps mass on several moves wherever more
than one is reasonable and sharpens only when the search is itself confident, as in forced or clearly
decided positions. Visit counts are thus high-entropy exactly where exploration matters and
low-entropy where it does not. The argument does not hinge on the specific search variant: more recent Lc0 training can use
Gumbel-MuZero \citep{danihelka2022gumbel}, which selects root actions by sampling without replacement
rather than by the visit counts above but still returns an \emph{improved} policy over several root
moves. Step~2 needs only that the exported target keep mass on more than one move where more than one
is reasonable, which the released network's policy does.

\paragraph{Step 2: distillation is mass-covering, so the network inherits the spread.}
The base network is not the search but a network trained to imitate it: its policy head is fit to the
visit-count targets by cross-entropy, which decomposes as
\[
  -\sum_{a} \pi_{\mathrm{mcts}}(a\mid s)\,\log\pibase(a\mid s)
  = \Ent\!\big(\pi_{\mathrm{mcts}}(\cdot\mid s)\big) + \KL{\pi_{\mathrm{mcts}}(\cdot\mid s)}{\pibase(\cdot\mid s)},
\]
that is, up to the target's own entropy, the \emph{forward} KL from the visit-count distribution to
the network. Forward KL is mass-covering (Section~\ref{subsec:structured},
Figure~\ref{fig:mechanism}): it penalizes dropping any move the search visited. A network that
collapsed onto one move would pay a large cross-entropy on every position where the search spread its
visits, so training transfers the breadth of the visit-count distributions onto the network's
single-pass policy. The mechanism is the one our method re-uses: the network was produced by a forward
KL toward exploratory MCTS targets, and we fine-tune it with a forward KL toward the network's own
resulting policy.

\paragraph{Consequence.}
The released network's single-pass policy $\pibase$, our frozen reference, is therefore exploratory by
construction: it carries forward the multi-move spread of the visit counts it was distilled from and
sharpens only where the search was decisive. The enabling condition of Section~\ref{subsec:structured}
holds for this base network without any tuning on our part, so the forward-KL penalty has genuine,
search-derived breadth to cover. At the opening position, for instance, the released network keeps its
probability spread across several first moves rather than committing to one. A
near-deterministic expert, such as a policy distilled from a single oracle move per position, would
offer no such breadth, and prior-directed exploration would then collapse toward that one move.

\section{Self-play data generation}
\label{app:selfplay}

Data are generated by running $N$ games in parallel and advancing each by one ply per training
iteration. At the start of training only, each slot is seeded with a random opening: we draw
$k\sim\mathrm{Uniform}\{0,1,\dots,60\}$ and play the first $k$ full moves by sampling from $\pibase$ (tempered by Equation~\eqref{eq:behavior} in the adaptive-temperature arms). This one-time
randomization staggers the games so the batch spans all phases of play. Thereafter a terminated game
restarts from the standard initial position, so that every restarted game is played by the agent
from the first move and the data reflect the agent's own state distribution rather than base-sampled
openings; because games end at different times, the batch stays phase-diverse on its own. Each
iteration forms a batch of one position per active game. At state $s$ we query the reference for
$\pibase(\cdot\mid s)$, the online model for $\pith(\cdot\mid s)$ and $v_\theta(s)$, and sample the
played move from $\mu(\cdot\mid s)$ of Equation~\eqref{eq:behavior}. Applying it gives $s'=T(s,a)$,
which the EMA model evaluates for the bootstrap value $v_{\bar\theta}(s')$ (the true one-hot result
when $s'$ is terminal). One gradient step is then taken, so every transition is used once while
fresh.

\section{Importance sampling, advantage, and the single-step update}
\label{app:objectives}

\paragraph{Truncated importance sampling.}
Because moves are drawn from the tempered $\mu$ rather than from $\pith$, the transitions are
off-policy with respect to the optimized policy. We correct this with a per-sample weight
\begin{equation}
  \rho(s,a) = \min\!\Big(1,\ \tfrac{\pith(a\mid s)}{\mu(a\mid s)}\Big),
  \label{eq:isweight}
\end{equation}
treated as a stop-gradient constant and applied to both the policy gradient and the value loss. It
corrects only the temperature-induced gap between $\mu$ and $\pith$; with one update per fresh batch
there is no policy staleness. Truncating at $1$ follows the V-trace estimator of the importance weighted actor-learner
architecture (IMPALA) \citep{espeholt2018impala}: the sharpened $\mu$ oversamples high-probability moves, which receive
$\rho<1$, while rarely sampled moves would otherwise carry large weights, and the cap bounds their
variance at the cost of a controlled bias. That bias sits exactly where $\mu$ sharpens most: in a
near-decided position an off-argmax sample has $\pith/\mu\gg1$ and is capped to $1$, but the
advantage there is itself near zero---the successor's WDL barely moves from the mover's view---so
the uncorrected mass carries almost no gradient. The truncation also reinforces the temperature's intent:
near-decided positions, where $\mu$ is sharp, are down-weighted to $\rho<1$ and terminate quickly,
leaving the update concentrated where $\mu\approx\pith$ and the advantage is informative.

\paragraph{Advantage and no normalization.}
The one-step TD advantage of Equation~\eqref{eq:advantage} pairs the EMA-evaluated successor value,
flipped to the mover's view, with the online estimate at $s$. TD(0) \citep{sutton1988learning} suits
this setting because the reference value head is already accurate, so a single bootstrap gives a
low-variance, low-bias target. We apply no advantage normalization, separating the two normalizations usually bundled together.
\emph{Mean-centering}: the value loss regresses $v_\theta(s)$ toward the same bootstrapped target that enters $A$, and the head starts near-consistent, so over the symmetric self-play population $\E[A]\approx0$ and a global baseline is near zero. This is an \emph{unconditional} statement that need not hold per batch or in transients,
where the EMA lag and any value-head miscalibration give the one-step residual a systematic sign; we
therefore omit per-batch mean-centering as a cheap option rather than rule it out. \emph{Std-scaling}:
dividing by a running standard deviation would rescale the natural units of the advantage, a
difference of win probabilities in $[-1,1]$, inflating the signal in quiet, low-variance positions
relative to sharp ones and injecting a position-dependent bias, so we deliberately avoid it.

\paragraph{Single-step reduction of PPO.}
We take one gradient step per fresh batch, which lets us drop the clipping that defines PPO
\citep{schulman2017ppo}. The clipped surrogate is
\begin{equation}
  \mathcal{L}^{\mathrm{PPO}}(\theta) = \E_{(s,a)}\!\left[\min\!\big(r(\theta)\hat A,\
  \clip(r(\theta),1-\epsilon,1+\epsilon)\hat A\big)\right], \quad r(\theta)=\tfrac{\pith(a\mid s)}{\piold(a\mid s)},
  \label{eq:ppo}
\end{equation}
with $\hat A(s,a)=\rho(s,a)A(s,a)$. The ratio $r$ is measured against the policy $\piold$ that
collected the data, and the clip enforces a trust region \citep{schulman2015trust} only when rollouts
are reused over several epochs, during which $\pith$ drifts from $\piold$. With a single update there
is no reuse: collection and differentiation parameters coincide, $r(\theta)\equiv1$, the two branches
of the minimum are identical, and the clip never activates. The surrogate reduces to the
advantage-weighted policy gradient
\begin{equation}
  \mathcal{L}^{\mathrm{PG}}(\theta) = \E_{(s,a)}\!\left[\rho(s,a)A(s,a)\log\pith(a\mid s)\right],
  \label{eq:pg}
\end{equation}
with $\rho$ and $A$ fixed, matching the single-update-per-batch practice of group relative policy optimization (GRPO)
\citep{shao2024deepseekmath}; the truncated weight is still required because $\mu$ differs from
$\pith$ through the temperature, only the trust-region clip is dropped; a variance-control device
thus survives, on the sampling ratio $\pith/\mu$ rather than the temporal ratio $r$, which never
leaves $1$. Dropping it removes the only
explicit step-size constraint, so the per-step change in $\pith$ is governed by the (small, layerwise)
learning rate rather than by a ratio bound---the EMA target smooths the value estimates that feed
$A$ but does not bound the policy step---the same trade GRPO-style single-epoch
updates make; we treat the single step as a simplification, not a stability guarantee in itself.

\paragraph{Value learning and the enumerability remark.}
The value head regresses its WDL prediction at $s$ onto the point-of-view-flipped EMA target
$\hat\wdl(s)=\operatorname{flip}(\wdl_{\bar\theta}(s'))$ (the one-hot result at terminal $s'$) by
cross-entropy,
\begin{equation}
  \mathcal{L}_V(\theta) = -\,\E_{(s,a)}\!\Big[\textstyle\sum_{o\in\{\mathrm{W},\mathrm{D},\mathrm{L}\}}
  \hat p^{\,o}(s)\log p^o_\theta(s)\Big],
  \label{eq:value-loss}
\end{equation}
also weighted by $\rho(s,a)$. This $\rho$-weighting is what makes the regression target
\emph{on-policy}. The value $v_\theta(s)$ should be the one-step
value of $s$ under the online policy, $\E_{a\sim\pith}[\,1-v_{\bar\theta}(s')\,]$, since that is the
baseline subtracted in the advantage of Equation~\eqref{eq:advantage}. Because moves are drawn from
the tempered $\mu$, the unweighted target $1-v_{\bar\theta}(s')$ averages to the child value under
$\mu$; weighting by $\rho(s,a)=\min(1,\pith/\mu)$ corrects the action over which the target is taken,
so in expectation the regression targets the $\pith$-value instead. The action-independence of
$v_\theta(s)$ does not reduce this to a state reweighting: the target
$\operatorname{flip}(\wdl_{\bar\theta}(s'))$ depends on the sampled move through $s'$, and $\rho$
corrects \emph{which child} it averages over. As on the policy gradient, the truncation at $1$ leaves
the usual V-trace bias (its fixed point sits between the $\mu$- and $\pith$-values for under-sampled,
high-ratio moves), and the single-step weight corrects the immediate child expectation but not the
state-visitation distribution---appropriate for a baseline fit on the states self-play visits. The
forward KL, finally, enters as an explicit penalty on the policy loss (as the entropy bonus it
replaces would), not as part of the return, so the critic estimates the reward-value of the current
policy---the consistent baseline---rather than a max-entropy-style soft value. Finally, on the cost of the forward penalty: the reverse KL
$\KL{\pith}{\pibase}$ is an expectation under $\pith$, estimated for free along the policy's own
rollouts from the reference's log-probability of each realized action
\citep{schulman2020kl}, whereas the forward KL
$\KL{\pibase}{\pith}$ needs the reference's full distribution at each state. For a language model
the exact sum runs over the vocabulary at every token; the sampled alternatives---drawing from the
reference offline \citep{li2025divergence} or importance-reweighting the policy's own rollouts
\citep{deng2025rapo}---avoid it, but the importance weights $\pibase/\pith$ are unbounded exactly
where the policy has dropped reference mass, the regime the penalty is meant to police. Here
the reference is an explicit categorical over the few legal moves from one network evaluation, so
Equation~\eqref{eq:kl} is computed exactly with no extra sampling.

\section{Strength evaluation protocol}
\label{app:metrics}

Searchless Elo is fit with BayesElo from a closed round-robin among the conditions---the frozen base
and the fine-tuned agents---all playing searchless: each move is the $\argmax$ of a single policy
forward pass. Openings are sampled from the UHO (Unbalanced Human Openings) book---slightly
unbalanced positions that lower the draw rate and sharpen resolution between closely matched
engines---and paired, each played from both colors; games are capped at $512$ plies. Every pair of
swept models plays $200$
games ($100$ paired openings), and each model additionally plays $1{,}000$ games ($500$ paired openings) against the base.
The fit fixes the base at the scale's zero; we place the ladder on an absolute axis by identifying that
zero with the Chessformer paper's rating of the same network, $2374\pm37$ \citep{monroe2026chessformer},
an external anchor whose $\pm37$ uncertainty shifts all ratings together and leaves within-pool
differences unchanged. The $\Delta$Elo intervals in Table~\ref{tab:main} ($\approx\!\pm8.3$) are the BayesElo $95\%$
intervals from these game counts. They reflect game-sampling variance for one fixed pair of
checkpoints per cell; we do not repeat the fine-tuning across seeds, so they do not capture
training-run variance, and given the small gains the ranking among the top cells should be read as
indicative rather than seed-stable. Because the rating is fit over the whole pool, a model's ladder
position and its direct head-to-head record against the base can diverge---some models that rate near
zero outscore the base directly, and vice versa---so Table~\ref{tab:main} reports the direct
record alongside the ladder. Finally, the $2374$ anchor is the
Chessformer paper's engine round-robin rating for this network, on a different scale from the human
online-pool (Lichess) Elo sometimes quoted for searchless networks, so the absolute value is
comparable only within engine-versus-engine play.

\section{Training configuration}
\label{app:config}

\paragraph{Optimization.}
Table~\ref{tab:hyperparameters} lists the settings; this appendix gives the reasoning behind them.
We use AdamW ($\beta_1=0.9$, $\beta_2=0.98$, weight decay $0$) with a $200$-step linear
warmup from zero to the per-group peak, held constant for the remaining $1{,}800$ steps, for $2{,}000$
total optimizer steps. We use layerwise learning rates: $3{\times}10^{-6}$ on the backbone,
$1{\times}10^{-5}$ on its final block, $1{\times}10^{-5}$ on the policy head, and $3{\times}10^{-5}$ on
the value head. The final transformer block is treated as part of the policy readout rather than a
generic backbone layer, because in Chessformer it is the block that resolves move-specific features for
the policy head \citep{lin2026tracing}; we therefore match its rate to the policy head's so the layer
producing the move logits adapts as fast as the head consuming them. The value head, unconstrained by
the forward-KL prior and fit by a low-bias EMA-stabilized bootstrap, takes the largest rate. The
value-loss coefficient is $c_v=0.2$, the EMA decay is $\alpha=0.995$, and the importance-sampling cap is
exactly $1$. We apply no gradient clipping and train in float32 precision; dropout and stochastic depth are off during
fine-tuning.

\paragraph{Self-play, data, and sweep.}
Each iteration consumes one transition from each of $N=1{,}024$ parallel games, so $2{,}000$ steps cover
$\approx2.05$M transitions. Games are capped at $512$ plies (scored as draws at the cap), with standard
chess termination otherwise. The headline forward-KL weight is $\beta=10^{-2}$, chosen exactly
as every other arm's was, on the $10{,}000$-puzzle suite (Table~\ref{tab:sweep}) and confirmed at
scale by Figure~\ref{fig:kl_sweep}; each baseline in Table~\ref{tab:main} is likewise reported at
its own swept-best coefficient. Searchless Elo
is the closed $\argmax$ round-robin of Appendix~\ref{app:metrics} ($200$ games per swept model pair
and $1{,}000$ games against the base). Coefficient selection uses the $10{,}000$-puzzle suite of \citet{ruoss2024amortized} (Table~\ref{tab:sweep}); the
reported accuracies use the uniformly sampled $100{,}000$-puzzle and the $\geq\!2400$-rated
$20{,}000$-puzzle Lichess suites (Table~\ref{tab:main}) and the mate-in-$N$ suites of
\citet{lin2026tracing}, with mate-in-5 restricted to exact five-move mates (Table~\ref{tab:mate}).
Every cell uses the same $2{,}000$-step, $1{,}024$-game budget, so compute is
matched across conditions.

\begin{table}[tbp]
\centering
\caption{Coefficient selection on the $10{,}000$-puzzle suite (multi-solution scoring): overall accuracy (\%) by exploration term, self-play
temperature, and coefficient $\beta$, with $95\%$ Wilson intervals. Bold marks each row's maximum,
the model carried into Table~\ref{tab:main}. For reference, the frozen base scores
$93.52\pm0.48$ and the unregularized runs $93.99\pm0.47$ (adaptive $\tau$) and $94.04\pm0.46$
($\tau\!\equiv\!1$).}
\label{tab:sweep}
\footnotesize
\setlength{\tabcolsep}{4.5pt}
\begin{tabular}{llccccc}
\toprule
 & & \multicolumn{5}{c}{Coefficient $\beta$} \\
\cmidrule(lr){3-7}
Exploration term & Self-play $\tau$ & $10^{-4}$ & $10^{-3}$ & $10^{-2}$ & $10^{-1}$ & $1$ \\
\midrule
Entropy bonus           & adaptive          & $94.42\pm0.45$ & $94.19\pm0.46$ & $\mathbf{94.70}\pm0.44$ & $94.07\pm0.46$ & $82.77\pm0.74$ \\
                        & $\tau\!\equiv\!1$ & $\mathbf{93.92}\pm0.47$ & $91.24\pm0.55$ & $91.59\pm0.54$ & $90.53\pm0.57$ & $83.13\pm0.73$ \\
\addlinespace
Reverse-KL anchor       & adaptive          & $94.30\pm0.45$ & $94.22\pm0.46$ & $\mathbf{94.57}\pm0.44$ & $94.47\pm0.45$ & $93.93\pm0.47$ \\
                        & $\tau\!\equiv\!1$ & $94.11\pm0.46$ & $94.38\pm0.45$ & $\mathbf{94.63}\pm0.44$ & $94.30\pm0.45$ & $93.86\pm0.47$ \\
\addlinespace
Forward-KL prior (ours) & adaptive          & $94.37\pm0.45$ & $94.54\pm0.45$ & $\mathbf{94.67}\pm0.44$ & $94.52\pm0.45$ & $93.80\pm0.47$ \\
                        & $\tau\!\equiv\!1$ & $94.26\pm0.46$ & $94.38\pm0.45$ & $\mathbf{94.59}\pm0.44$ & $94.42\pm0.45$ & $93.78\pm0.47$ \\
\bottomrule
\end{tabular}
\end{table}

\section{Additional results}
\label{app:extra}

\begin{figure}[tbp]
\centering
\includegraphics[width=0.95\linewidth]{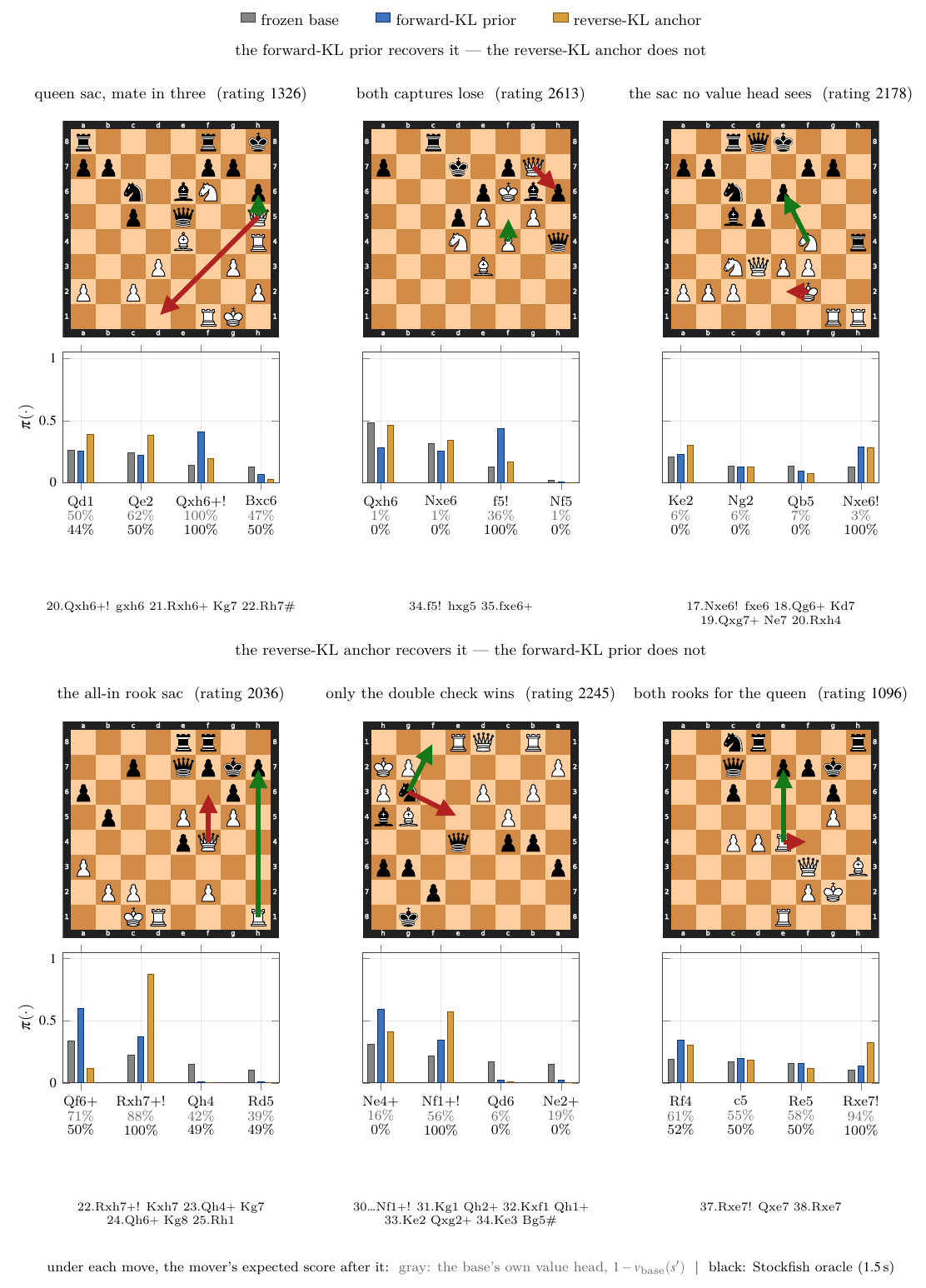}
\caption{Six exclusive recoveries from the $100{,}000$-puzzle suite---\textbf{top row:} puzzles
the forward prior newly solves and the reverse anchor does not; \textbf{bottom row:} the mirror.
Each panel shows the first solver move (side to move at the bottom): green arrow, the winning
move; red arrow, the move kept by the base and by the fine-tuned model that fails; bars, the three
models' probabilities of the labeled moves; under each move, the mover's expected score after
it---gray from the base's \emph{own} value head $1-v_{\mathrm{base}}(s')$, black from a
$1.5$-second Stockfish oracle; beneath each panel, the solution. Left to right: a queen sacrifice
into a three-move mate the base's value head already scores as certain (rating $1326$); an endgame
where the base's two favorite captures both lose and only the quiet 34.f5 wins ($2613$); a knight
sacrifice no value head sees ($2178$); a rook sacrifice the recovering model commits to at $0.88$
($2036$); knight discoveries of which only the double check wins ($2245$); both rooks for the
queen, the solution ranked fifth ($1096$).}
\label{fig:examples}
\end{figure}

\paragraph{Example exclusive recoveries.}
The two anchors' exclusive recoveries differ in kind (Section~\ref{subsec:recovery});
Figure~\ref{fig:examples} shows three of each, annotating every candidate move with
two verdicts on the position it leads to: the base's \emph{own} evaluation
$1-v_{\mathrm{base}}(s')$---the bootstrap quantity of Equation~\eqref{eq:advantage}---and a
$1.5$-second Stockfish oracle. In five of the six positions the base's verdict already separates
the winning move from the played one by $17$--$50$ percentage points, in the direction the
oracle confirms---clearest in the queen-sacrifice panel (top left), where the base's value head
scores the sacrifice $100\%$, certain of the forced mate, against at most $62\%$ for the
alternatives, yet its policy ranks it third: the
base's value head knows what its policy is failing to play, so the first
time self-play samples the retained move, the advantage signal is immediate. In the sixth (top
right) the oracle splits the candidates $100\%$ against $0\%$ and the base's value head
cannot---it scores the winning knight sacrifice \emph{lowest}, $3\%$ against
$6$--$7\%$---there the repair is pure policy re-ranking. The misses are instructive
in both directions. The reverse anchor loses that same recovery by a photo finish ($0.302$
against $0.282$) and doubles down on another panel's queen retreat, letting the win slip
($0.26\to0.39$ on 20.Qd1?); the forward prior doubles down
on 30\ldots Ne4$+$?, the discovered check that loses outright where only
the double check mates ($0.31\to0.60$), and on the drawing 22.Qf6$+$? ($0.34\to0.60$), a check
the base's value head itself misreads as winning ($71\%$). Sharpening a near miss the wrong way
is a risk neither anchor escapes. And the
bottom-right panel keeps the aggregate honest: the deepest recoveries in the study sit at
rank~$5$, and this one---both rooks for the queen, ranked fifth at $\pi=0.100$, which the base's
value head already scores at $94\%$ against $61\%$ for the base's own choice---belongs to the
reverse anchor.

\begin{figure}[tbp]
\centering
\includegraphics[width=\linewidth]{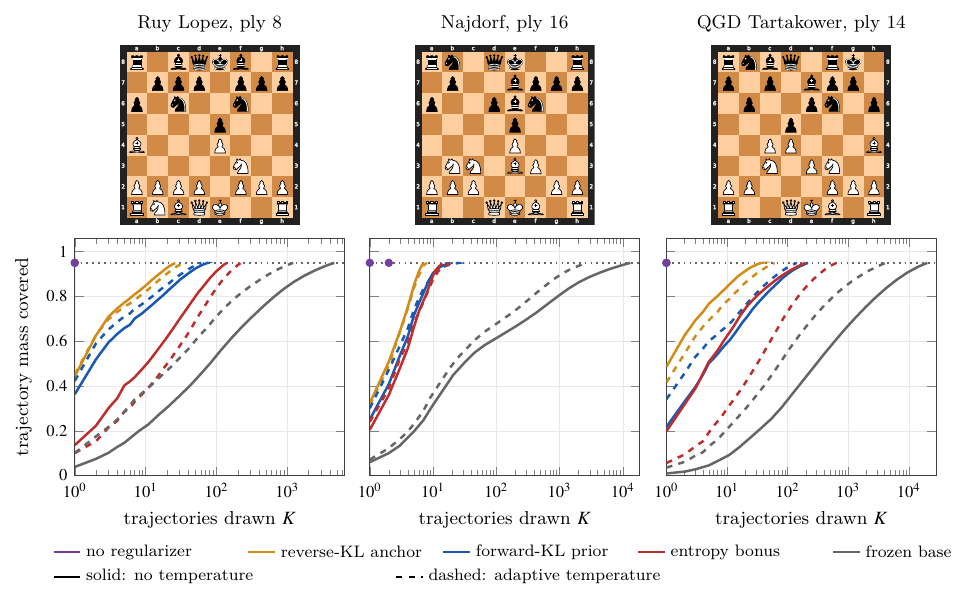}
\caption{Trajectory coverage $\Omega^{N=4}_K$ beyond the initial position, under the protocol of
Figure~\ref{fig:omega}: cumulative depth-$4$ trajectory mass of the first $K$ draws without
replacement, mean over $100$ sampling seeds; every curve ends at the $95\%$ extraction target
(dotted). Solid lines are the no-temperature configurations (the frozen base sampled raw),
dashed their adaptive-temperature counterparts (the base sampled through the tempered $\mu$);
the dot marks the unregularized configurations, whose first draw covers $95\%$. Each position is
shown above its panel, full move history
supplied (all White to move): \textbf{left}, the Ruy Lopez mainline (ply $8$, $27$ legal moves,
$\approx\!7.0\times10^5$ depth-$4$ continuations); \textbf{middle}, the Najdorf Sicilian English
Attack (ply $16$, $47$ legal moves, $\approx\!2.6\times10^6$); \textbf{right}, the Queen's Gambit
Declined Tartakower (ply $14$, $38$ legal moves, $\approx\!1.5\times10^6$).}
\label{fig:omega-positions}
\end{figure}

\paragraph{Coverage beyond the initial position.}
Figure~\ref{fig:omega-positions} repeats the measurement of Figure~\ref{fig:omega} at three
opening tabiyas of different character---the Ruy Lopez mainline (open, classical), the Najdorf
Sicilian English Attack (sharp, asymmetric), and the Queen's Gambit Declined (QGD) Tartakower (closed,
positional)---and the initial-position picture reproduces at all three. The frozen base stays
widest by more than an order of magnitude, covering $95\%$ of its depth-$4$ mass in $4{,}599$,
$13{,}149$, and $20{,}247$ draws raw ($1{,}211$, $2{,}773$, and $4{,}145$ through the tempered
$\mu$), while the unregularized policy collapses to a single line in every position. The
mechanism's central prediction holds throughout: the mode-seeking reverse anchor is the most
concentrated of the exploration terms---$95\%$ coverage in $26$--$34$, $7$--$8$, and $36$--$61$
draws respectively---always tighter than the mass-covering forward prior ($61$--$73$, $19$--$24$,
and $142$--$217$). The forward prior and the entropy bonus both fall between the reverse anchor
and the base, and which of the two is broader is not fixed---the entropy bonus in the Ruy Lopez,
the forward prior in the Najdorf, and split by temperature variant in the QGD---but their common
separation from the tightly concentrated reverse anchor below and the sprawling base above is
unchanged.

\end{document}